\documentclass{article}
\PassOptionsToPackage{numbers, compress}{natbib}
\usepackage[preprint]{neurips_2026}

\usepackage[utf8]{inputenc}
\usepackage[T1]{fontenc}
\usepackage{hyperref}
\usepackage{url}
\usepackage{booktabs}
\usepackage{amsfonts}
\usepackage{amssymb}
\usepackage{amsmath}
\usepackage{nicefrac}
\usepackage{microtype}
\usepackage{xcolor}
\usepackage{graphicx}
\usepackage{subcaption}
\usepackage{adjustbox}
\usepackage{tikz}
\usetikzlibrary{positioning,arrows.meta}
\usepackage{algorithm}
\usepackage{algpseudocode}
\usepackage{xspace}
\usepackage{placeins}
\usepackage{enumitem}
\usetikzlibrary{positioning,arrows.meta,calc,fit,shapes.geometric}
\usepackage{graphicx}
\usepackage{amsmath,amssymb}
\usepackage{xcolor}

\usepackage{wrapfig}

\newcommand{\SkillMigrator}{\textsc{SkillMigrator}\xspace}

\newcommand{\TIP}{TIP\xspace}

\title{Beyond Domains: Reusing Web Skills via Transferable Interaction Patterns}

\author{%
  \bfseries
  Shiqi He\textsuperscript{1}\thanks{Correspondence to: \texttt{shiqihe@umich.edu}}, Yue Cui\textsuperscript{2}, Feijie Wu\textsuperscript{3}, Xinyu Ma\textsuperscript{4}, \\
  \bfseries
  Jiaheng Lu\textsuperscript{5}, Yaliang Li\textsuperscript{2}, Bolin Ding\textsuperscript{2}, Mosharaf Chowdhury\textsuperscript{1} \\[0.6ex]
  \normalfont\small
  \textsuperscript{1}University of Michigan \quad
  \textsuperscript{2}Alibaba Group \quad
  \textsuperscript{3}Purdue University \\
  \normalfont\small
  \textsuperscript{4}McMaster University \quad
  \textsuperscript{5}University of Pennsylvania
}
\usepackage{ifthen}
\usepackage[normalem]{ulem} 
\usepackage{xcolor}
\usepackage{amssymb}

\newboolean{showedits}
\setboolean{showedits}{false} 
\ifthenelse{\boolean{showedits}}
{
	\newcommand{\del}[1]{\textcolor{red}{\sout{#1}}} 
}{
	\newcommand{\del}[1]{} 
	
}

\newboolean{showcomments}
\setboolean{showcomments}{false} 
\newcommand{\id}[1]{$-$Id: scgPaper.tex 32478 2010-04-29 09:11:32Z oscar $-$}

\ifthenelse{\boolean{showcomments}}
{\newcommand{\nbc}[3]{
 {\colorbox{#3}{\bfseries\sffamily\scriptsize\textcolor{white}{#1}}}
 {\textcolor{#3}{\sf\small$\blacktriangleright$\textit{#2}$\blacktriangleleft$}}}
 }
{\newcommand{\nbc}[3]{}
 \renewcommand{\del}[1]{} 
 }

\definecolor{ibcolor}{rgb}{0.9,0.5,0}
\definecolor{dsrcolor}{rgb}{0.5,0.6,0}
\definecolor{cfcolor}{rgb}{0,0.5,0.9}
\definecolor{lwcolor}{rgb}{0.2,0.8,0.4}
\definecolor{eycolor}{rgb}{0.7,0.6,1.0}
\definecolor{oldcolor}{rgb}{0.2,0.2,0.2}
\definecolor{tdcolor}{rgb}{0.0,0.5,0.7}

\begin{document}

\maketitle
\setcounter{footnote}{0}

\begin{abstract}
Large language model (LLM) web agents are usually deployed as
\emph{tool callers}: each turn, the model reads a fresh page
observation and emits one structured tool action.  When every
action is a low-level primitive, horizons grow quickly and so do
policy-facing LLM completions, dominating latency and cost on
benchmarks such as Mind2Web and WebArena.  Recent systems
therefore wrap repeated interaction fragments as \emph{web
skills}: callable tools built from successful trajectories or
induced programs, so one call can replace several primitives.  
However, prior skill libraries are still triggered
mainly by instruction similarity or coarse site metadata, which
yields low \emph{skill reuse} on held-out sites and leaves much of
the potential step and token reduction on the table.

We present \SkillMigrator{}, an agent that learns reusable web skills and transfers them across sites by matching layout structure rather than specific element references.
Each induced skill is stored as a \emph{transferable interaction pattern} (\TIP{}): the skill paired with a structural sketch of the snapshot at induction time.
At test time, \SkillMigrator{} retrieves \TIP{}s by layout similarity and grounds their references on the live page.
The rest of the stack is standard: accessibility-snapshot observations with stable references, and fixed tool calling over primitives plus skill invocations.
Compared with the state-of-the-art approaches, \SkillMigrator{} reduces the average LLM-action count on successful trajectories by 8–10\% across both WebArena and Mind2Web at matched success rate.
\end{abstract}

\section{Introduction}
\label{sec:introduction}

Web agents translate a user's natural-language goal into a sequence of browser actions such as searching, clicking, typing, and submitting forms, providing a general interface for automating tasks that are difficult to script manually.
Recent benchmarks such as WebShop \cite{yao2022webshop}, Mind2Web \cite{deng2023mind2web}, and WebArena \cite{zhou2024webarena} cover e-commerce interaction, open-domain websites, and realistic self-hosted environments, highlighting both the practical value and the difficulty of this setting.
However, most existing web agents rely on an LLM-centered decision loop that repeatedly queries the LLM to predict the next action from the current webpage state, often following reasoning-and-acting paradigms such as ReAct \citep{yao2023react}.
This design is flexible but expensive at deployment, since each task may require many sequential LLM calls, with cost and latency growing in the length and number of interaction trajectories \citep{yang2024agentoccam}.
There is therefore a need for a \emph{cost-effective web agent} that reduces reliance on the LLM.
%

\definecolor{ecBlue}{HTML}{2E5AAC}
\definecolor{devGreen}{HTML}{2D7A4F}
\definecolor{forumRed}{HTML}{B53A3A}
\definecolor{skillOrange}{HTML}{D97706}
\definecolor{actBg}{HTML}{F4F7FB}
\definecolor{actBorder}{HTML}{B6C4DC}
\definecolor{skillBg}{HTML}{FFF4E0}
\definecolor{skillBorder}{HTML}{D97706}
\definecolor{slotTitle}{HTML}{1F77B4}   
\definecolor{slotBody}{HTML}{2CA02C}    
\definecolor{slotSubmit}{HTML}{D62728}  
 
\begin{figure}[t]
  \centering
  \begin{tikzpicture}[
      font=\footnotesize,
      every node/.style={align=center, inner sep=1pt},
      domLab/.style ={font=\bfseries\footnotesize},
      taskLab/.style={font=\itshape\scriptsize, text=black!72},
      imgBox/.style ={draw=black!35, line width=0.4pt,
                      rounded corners=1.5pt, inner sep=0.6pt},
      actBox/.style ={draw=actBorder, fill=actBg,
                      line width=0.4pt, rounded corners=2.5pt,
                      inner xsep=5pt, inner ysep=4pt,
                      align=left,
                      text width=3.55cm,
                      font=\ttfamily\scriptsize},
      skillBox/.style={draw=skillBorder, line width=0.7pt,
                       fill=skillBg, rounded corners=4pt,
                       inner xsep=10pt, inner ysep=6pt,
                       align=center,
                       text width=12.7cm,
                       font=\footnotesize},
      arr/.style    ={-{Latex[length=1.7mm]}, line width=0.65pt,
                      draw=skillOrange},
  ]
 
    \def\colW{4.0cm}
    \def\colSep{0.35cm}
 
    \node[imgBox] (img1) at (0,0)
      {\includegraphics[width=\colW]{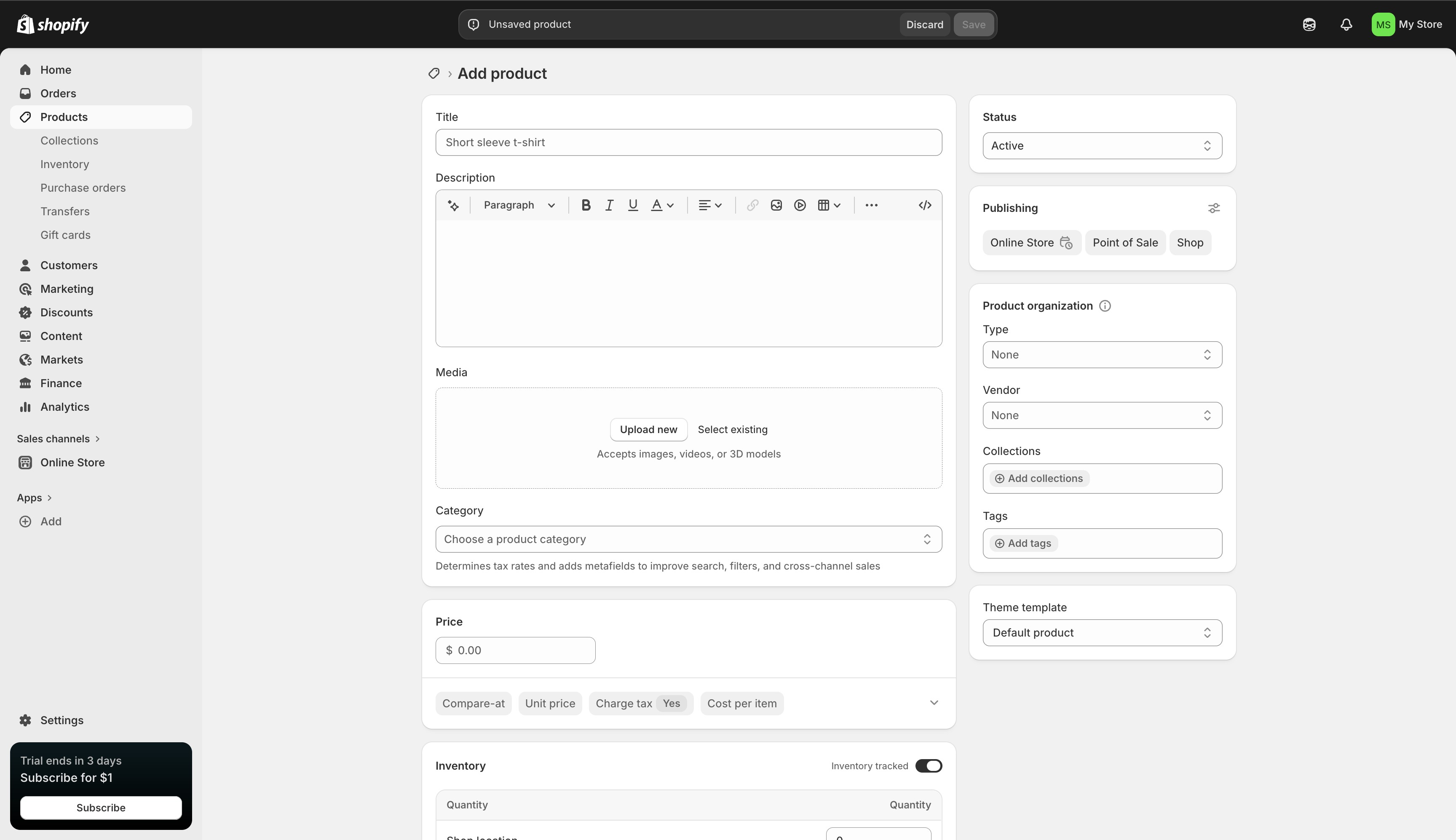}};
    \node[imgBox, right=\colSep of img1] (img2)
      {\includegraphics[width=\colW]{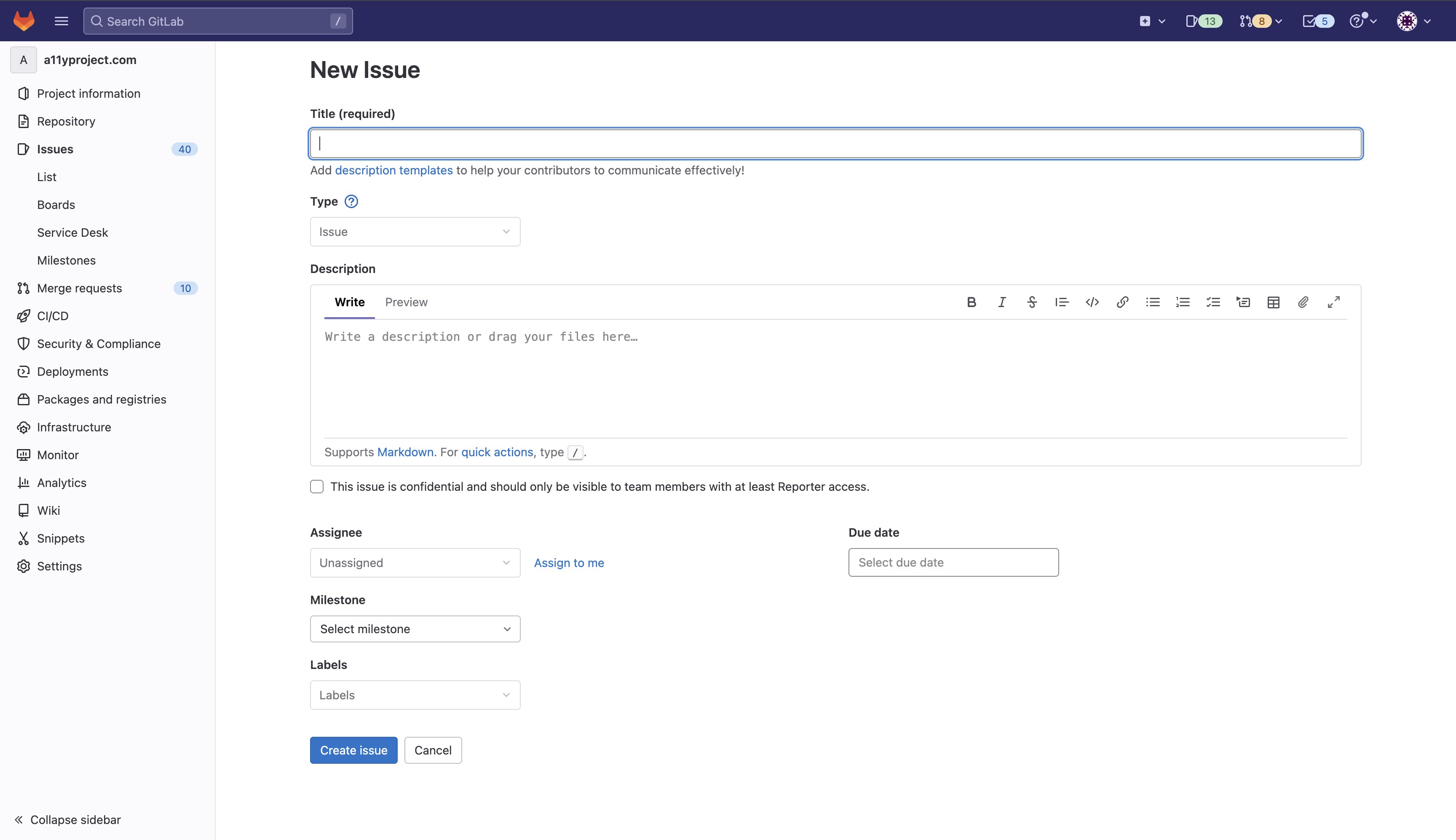}};
    \node[imgBox, right=\colSep of img2] (img3)
      {\includegraphics[width=\colW]{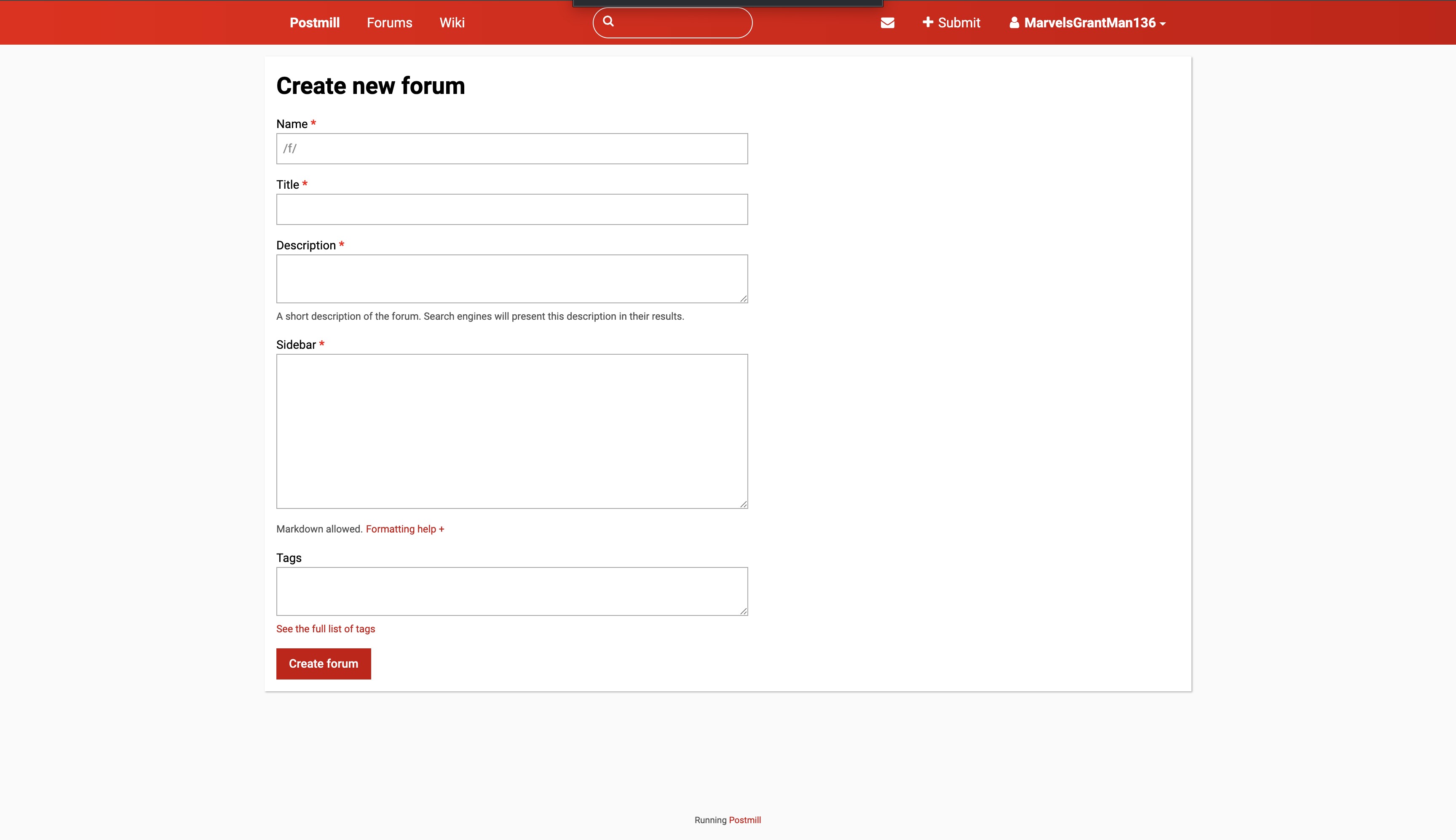}};
 
    \node[domLab, text=ecBlue,    above=2pt of img1.north]
          {Shopify\, $\bullet$\, E-commerce};
    \node[domLab, text=devGreen,  above=2pt of img2.north]
          {GitLab\, $\bullet$\, Developer Tools};
    \node[domLab, text=forumRed,  above=2pt of img3.north]
          {Postmill\, $\bullet$\, Online Forum};
 
    \node[taskLab, below=3pt of img1.south] (t1)
          {\emph{Task:} ``Add a new product''};
    \node[taskLab, below=3pt of img2.south] (t2)
          {\emph{Task:} ``Open a new issue''};
    \node[taskLab, below=3pt of img3.south] (t3)
          {\emph{Task:} ``Create a new forum''};
 
    \node[actBox, below=3pt of t1.south] (a1) {%
      \textbf{\textsf{\scriptsize Primitive actions}}\\[1pt]
      fill(\textcolor{slotTitle}{`Title'}, \ldots)\\
      fill(\textcolor{slotBody}{`Description'}, \ldots)\\
      fill(`Price', \ldots)\\
      click(\textcolor{slotSubmit}{`Save'})%
    };
    \node[actBox, below=3pt of t2.south] (a2) {%
      \textbf{\textsf{\scriptsize Primitive actions}}\\[1pt]
      fill(\textcolor{slotTitle}{`Title'}, \ldots)\\
      fill(\textcolor{slotBody}{`Description'}, \ldots)\\
      select\_option(`Type', \ldots)\\
      click(\textcolor{slotSubmit}{`Create issue'})%
    };
    \node[actBox, below=3pt of t3.south] (a3) {%
      \textbf{\textsf{\scriptsize Primitive actions}}\\[1pt]
      fill(`Name', \ldots)\\
      fill(\textcolor{slotTitle}{`Title'}, \ldots)\\
      fill(\textcolor{slotBody}{`Description'}, \ldots)\\
      click(\textcolor{slotSubmit}{`Create forum'})%
    };
 
    \node[skillBox, below=0.95cm of a2.south] (s) {%
      \textbf{One \TIP{} reusable across all three domains:}\\[2pt]
      $\iota$\,:\;\emph{``fill a labelled form and click submit''}\quad
      $\sigma$\,:\;\mbox{\emph{fill-and-submit} template}\quad
      $\Phi$\,:\;$\bigl\{\,$%
         \textcolor{slotTitle}{\emph{title}-like}, \;
         \textcolor{slotBody}{\emph{body}-like}, \;
         $\dots\,\bigr\}$\\[1pt]
      \textsf{plan} $\;\Rightarrow\;$
        \texttt{fill}$^{\ast}$ then \texttt{click}(\textcolor{slotSubmit}{\emph{submit}})
        \quad{\color{black!55}\scriptsize (replaces $n$ policy LLM calls with 1 skill call)}%
    };
 
    \coordinate (m) at ($(a2.south) + (0,-0.42)$);
    \draw[arr, line width=0.55pt] (a1.south)
      to[out=-90, in=170, looseness=0.85] (m);
    \draw[arr, line width=0.55pt] (a2.south) -- (m);
    \draw[arr, line width=0.55pt] (a3.south)
      to[out=-90, in=10,  looseness=0.85] (m);
    \draw[arr, line width=0.9pt] (m) -- (s.north);
 
  \end{tikzpicture}
  \caption{\textbf{Cross-domain skill reuse motivates \SkillMigrator{}.}
  Three websites drawn from very different
  domains---\emph{Shopify} (e-commerce), \emph{GitLab} (developer
  tools), and \emph{Postmill} (online forum)---use different page
  layouts, field vocabularies, and submit-button labels.  Yet the three
  subtasks reduce to the same programmatic pattern: fill a few labelled
  inputs, then click a single submit button.  Same-colour fields
  (\textcolor{slotTitle}{\emph{title}-like},
  \textcolor{slotBody}{\emph{body}-like},
  \textcolor{slotSubmit}{\emph{submit}}) are paraphrases of the same
  abstract slot across all three sites.  
  \SkillMigrator{} stores
  \emph{one} \TIP{}---intent~$\iota$, operation template~$\sigma$, slot
  schema~$\Phi$, and the induction-time tree skeleton~$\tau$---and
  reuses it on all three pages, replacing many policy LLM steps with a
  single skill call.}
  \vspace{-10pt}
  \label{fig:cross-domain-overview}
\end{figure}
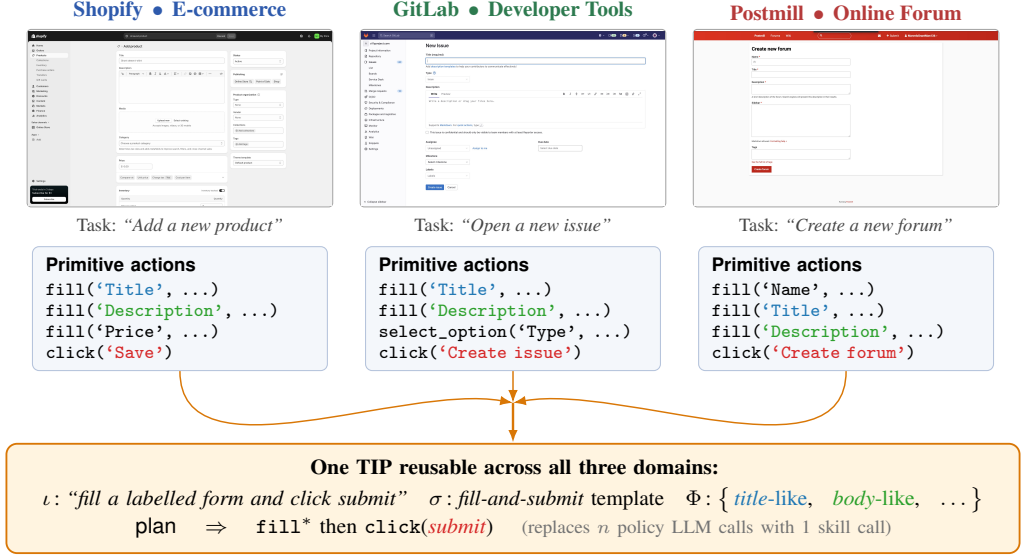


Recent work pursues this goal through reusable web skills, which store procedural knowledge from prior web interactions and reapply it in future tasks \citep{wang2024agent,wang2025inducing,zheng2025skillweaver,prabhu2025walt,wang2026webxskill}. 
A successful interaction trajectory is abstracted into a skill that the agent can retrieve and execute when it encounters a similar goal or webpage state.
This replaces many primitive LLM decisions with a single higher-level operation, reducing LLM calls and shortening the interaction trajectory.
Because the skill encodes a validated action pattern, it also mitigates compounding errors in long-horizon navigation \citep{wang2025inducing,zheng2025skillweaver,prabhu2025walt,lu2026contractskill}.

Existing skill-reuse methods fall into two types: reusing web skills under \emph{the same website}, where the skill is specialized to a specific site and its interface \citep{zheng2025skillweaver,prabhu2025walt,wang2026webxskill}, and reusing web skills \emph{under the same domain}, where the skill is transferred across websites that share similar task structures, such as shopping, maps, forums, or code repositories \citep{wang2024agent,wang2025inducing,yu2026polyskill}.
Both directions have important limitations.
Same-website methods suffer from low reuse rates, because each skill is tied to the specific interface, DOM structure, and interaction pattern of the site where it was learned \citep{zheng2025skillweaver,prabhu2025walt,wang2026webxskill}.
The agent benefits only when future tasks revisit that site, which is restrictive in open-web settings where user requests span diverse websites.
Same-domain methods such as PolySkill \citep{yu2026polyskill} address this through polymorphic abstractions that separate shared skill interfaces from site-specific implementations, enabling reuse across sites within a domain.
However, they still confine reuse to a single domain and miss the fact that sites across different domains often share strong interaction patterns, as illustrated in Figure~\ref{fig:cross-domain-overview}.
As a result, current skill-reuse methods remain narrower than necessary, motivating a more general form of reusable web skills that can transfer beyond both \emph{the same website} and \emph{the same domain}.

Obtaining reusable web skills beyond both {the same website} and {the same domain} is challenging because of unreliable skill retrieval. 
Existing methods retrieve candidates using instruction similarity, intent labels, or website metadata \citep{wang2024agent,zheng2025skillweaver,yu2026polyskill}.
However, these signals are insufficient for cross-domain transfer: two tasks with different wording may require the same interaction program, while two textually similar tasks may require different DOM-level control flow. 
As a result, the agent may fail to retrieve useful skills or mistakenly execute unsuitable ones, forcing it to fall back to primitive action generation with frequent LLM calls.

We propose \SkillMigrator{}, a cost-effective web agent that enables skill reuse beyond both \emph{the same website} and \emph{the same domain}. \SkillMigrator{} follows the standard programmatic-skill setting, where observations are accessibility snapshots with stable element references and actions are issued through a fixed tool-calling API over primitive actions and skill invocations \citep{zheng2025skillweaver,prabhu2025walt,yu2026polyskill}. 
Its memory unit is a transferable interaction pattern (\TIP{}) , which pairs each induced skill with a structural sketch of the webpage snapshot where the skill was validated. 
At inference time, \SkillMigrator{} retrieves skills from a single global library by combining layout similarity with text signals, then grounds the matched abstract constraints to live element references before replaying the skill. 
This design allows the agent to identify reusable interaction patterns across websites with different wording, interfaces, and domains, while avoiding the execution of weakly matched skills.


%

%

%


\paragraph{Contributions.}
%
\begin{itemize}[topsep=0pt,itemsep=0pt,parsep=0pt,partopsep=0pt,leftmargin=*]
    \item To our knowledge, this is the first work to study reusable web skills across websites beyond domains. 
        This setting is non-trivial because similar interaction patterns may appear with different layouts, labels, and DOM structures. 
    \item We propose \SkillMigrator{} for cross-domain skill matching.
        It stores induced skills as TIPs, each pairing a validated skill with the structural sketch of its source webpage. 
        For a new task, \SkillMigrator{} retrieves relevant TIPs using layout and text signals, grounds them to live webpage elements, and falls back to primitive control when no reliable match is found. 
    \item We empirically compare \SkillMigrator{} against existing web-agent baselines on Mind2Web and WebArena, reducing average LLM-action count on successful trajectories by 8--10\% relative to the state-of-the-art baselines at matched task success rate.
\end{itemize}



\section{Background and Motivation}
\label{sec:background}

\subsection{Preliminaries and Problem Formulation}

\begin{wraptable}{r}{0.55\linewidth}
  \centering
  \footnotesize
  \vspace{-15pt}
  \caption{Representative primitive tools. $e$ is an element ref in the text snapshot. Skill calls expand into grounded primitive calls.}
  \vspace{-5pt}
  \label{tab:action-space}
  \resizebox{\linewidth}{!}{%
  \begin{tabular}{@{}ll@{}}
    \toprule
    Action & Description \\
    \midrule
    \texttt{noop} & Wait or synchronize. \\
    \texttt{click(}$e$\texttt{)} & Activate control $e$. \\
    \texttt{fill(}$e$, \textit{text}\texttt{)} & Type into an input. \\
    \texttt{scroll(}$e$, \textit{dir}\texttt{)} & Scroll element or viewport. \\
    \texttt{select\_option(}$e$, \textit{value}\texttt{)} & Choose an option. \\
    \texttt{tab\_focus(}\textit{index}\texttt{)} & Switch browser tab. \\
    \texttt{go\_back} / \texttt{go\_forward} & History navigation. \\
    \texttt{send\_msg\_to\_user(}\textit{text}\texttt{)} & Final user-visible answer. \\
    \texttt{stop} & Terminate with success or failure. \\
    \midrule
    \texttt{call\_skill(}$\sigma$\texttt{,}\,\textit{args}\texttt{)} &
    \parbox[t]{0.35\linewidth}{Invoke skill $\sigma$ and execute its implementation.} \\
    \bottomrule
  \end{tabular}}
  \vspace{-10pt}
\end{wraptable}
 
\paragraph{Web Agent Environment.}
We follow the BrowserGym and WebArena
convention~\citep{drouin2024workarena,zhou2024webarena}: at each
timestep $t$, the agent receives a Playwright-style accessibility
snapshot $o_t$ with stable refs, roles, names, and state attributes,
and emits one tool call $a_t \sim \pi_\theta(\cdot \mid q, o_{0:t},
a_{0:t-1})$ from the action space $\mathcal{A}$ in
Table~\ref{tab:action-space}.  For benchmarks, we consider
Mind2Web~\citep{deng2023mind2web} (cross-task, cross-website,
cross-domain splits) and WebArena~\citep{zhou2024webarena}
(812 executable tasks across shopping, admin, reddit, gitlab, map,
multisite).  The default policy input is text-only and
screenshot-grounded or pixel-space agents are outside our
scope~\citep{he2024webvoyager,koh2024visualwebarena}.  A full formulation is in Appendix~\ref{app:formulation}.

\paragraph{Skill library.}
Beyond primitive browser actions, recent web agents~\citep{wang2025inducing, zheng2025skillweaver, yu2026polyskill} are equipped with a
skill library $\mathcal{K}$.  Each skill $k\in\mathcal{K}$ is a
temporally extended routine~\citep{sutton1999between} that maps a
subtask $s$ and observation $o$ to a short action sequence $k(s,o)
= \langle\tilde{a}_1,\dots,\tilde{a}_n\rangle$ over
Table~\ref{tab:action-space}, exposed to the policy as a callable
high-level macro, which reuses recurring interaction patterns such as
opening a menu, filling a form, searching, or
filtering.


\paragraph{Problem Formulation.}
Given an instruction $q$, a planner decomposes it into a sequence
of subtasks $\mathbf{s}(q)=\{s_1,\ldots,s_{T_q}\}$.
Let $\tilde{o}_0$ be the initial observation and $\tilde{o}_i$
the observation after finishing $s_i$.  We define
$\mathcal{N}(s,o\mid\mathcal{K},\pi_\theta)$ as the number of
\emph{new} primitive actions emitted by $\pi_\theta$ to complete
$s$ from $o$: if the subtask is fully covered by a retrieved
skill, $\mathcal{N}=0$; otherwise the agent falls back to
$\pi_\theta$ and $\mathcal{N}=n$ where $n$ is the length of the
fallback trajectory.  Our goal is to construct a compact skill
library that minimises the expected number of LLM-generated
primitive actions across tasks:
\begin{equation}
    \min_{\mathcal{K}} 
    \mathbb{E}_{q \sim Q}
    \left[
    \sum_{i=1}^{T_q}
    \mathcal{N}(s_i,\tilde{o}_{i-1} \mid \mathcal{K},\pi_\theta)
    \right]
    + \lambda\, C(\mathcal{K}),
    \label{eq:problem}
\end{equation}
where $Q$ is the task distribution, $C(\mathcal{K})$ is the
library cost, and $\lambda$ trades off action savings against
library size.

\subsection{Skill Induction Methods}

\begin{figure}[t]
  \centering
  \begin{subfigure}[t]{0.47\linewidth}
    \centering
    \includegraphics[width=\linewidth]{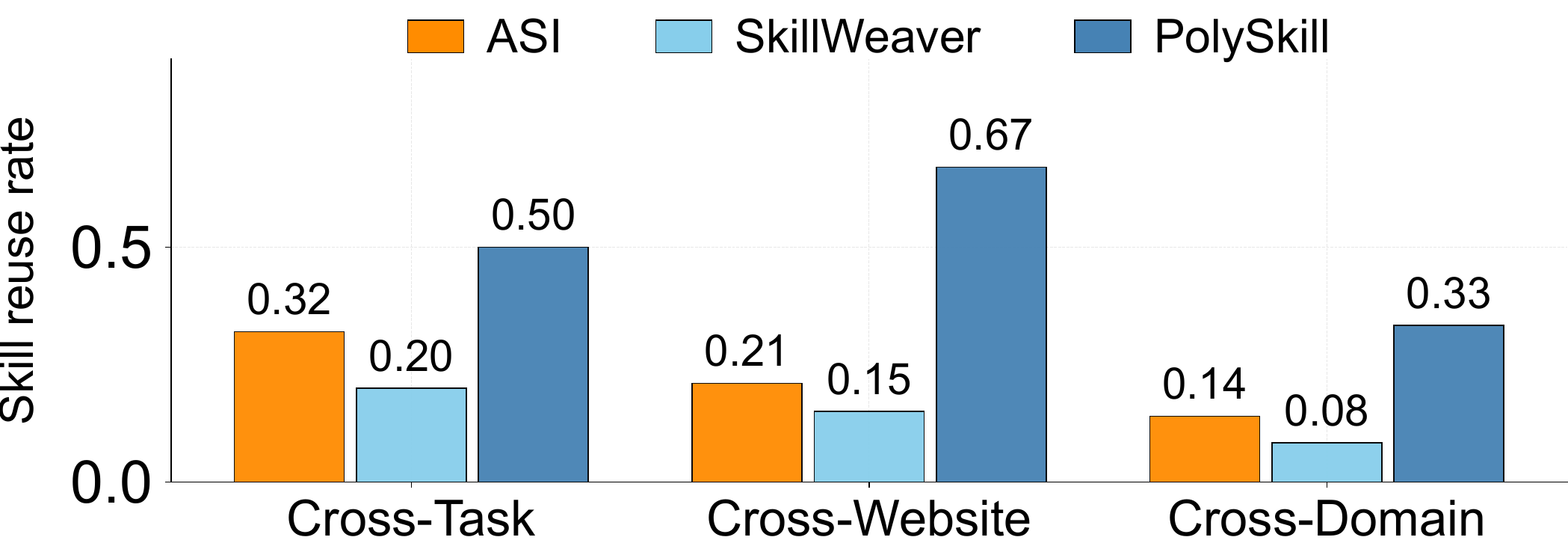}
    \caption{Skill reuse rate.}
    \label{fig:reuse-motivation-reuse}
  \end{subfigure}
  \hfill
  \begin{subfigure}[t]{0.47\linewidth}
    \centering
    \includegraphics[width=\linewidth]{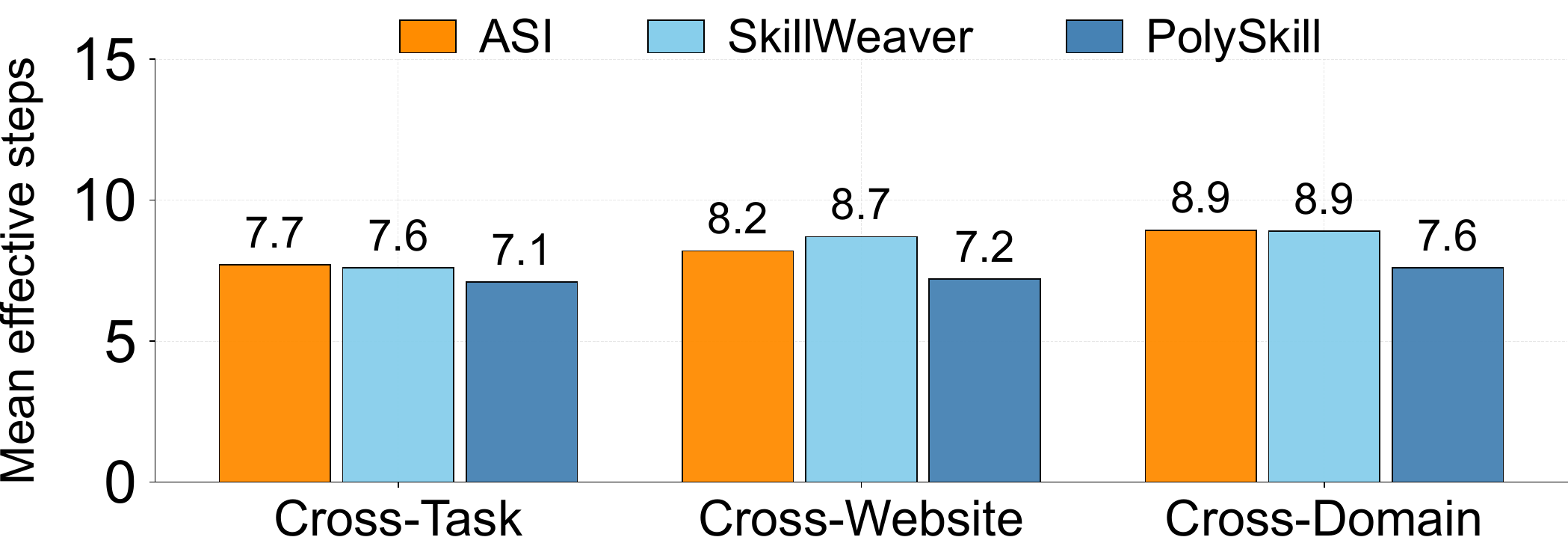}
    \caption{Mean effective steps.}
    \label{fig:reuse-motivation-steps}
  \end{subfigure}
  \hfill
  \caption{Motivating comparison of ASI, SkillWeaver, and PolySkill on a cumulative Mind2Web-style setup: cross-task (same website), cross-website (same domain, new sites), and cross-domain (new domains). Subfigure \subref{fig:reuse-motivation-reuse} reports per-phase skill reuse and \subref{fig:reuse-motivation-steps} show successful-step cost.}
  \label{fig:reuse-motivation}
\end{figure}

\paragraph{Existing Work.}
Recent work on reusable procedural knowledge for web and tool-use
agents includes textual workflow memory
(AWM~\citep{wang2024agent}), verified programmatic skills callable
as high-level actions (ASI~\citep{wang2025inducing}), self-induced
skill APIs from exploration
(SkillWeaver~\citep{zheng2025skillweaver}), polymorphic
cross-site abstractions (PolySkill~\citep{yu2026polyskill}), and
discovered website tools (WALT~\citep{prabhu2025walt}).  Together,
these methods compress repeated primitive sequences into reusable
abstractions, shortening interaction horizons and reducing policy
LLM calls.
 
\paragraph{Limitations.}
Despite these advances, skill retrieval remains a bottleneck on
unseen websites because existing methods rely on semantic keys like task descriptions, workflow summaries, skill names and API
descriptions.  Web tasks often preserve interaction structure
while changing surface wording, so purely semantic retrieval
under-retrieves reusable skills---raising
$\mathcal{N}(s,o\mid\mathcal{K},\pi_\theta)$---and over-retrieves
skills whose execution context is incompatible with the current
page.  Figure~\ref{fig:reuse-motivation} illustrates this on a
60-task Mind2Web subset evaluated with GPT-4.1: skill reuse drops
sharply for ASI, SkillWeaver, and PolySkill as the test trace moves
from cross-task to cross-website to cross-domain
(Figure~\ref{fig:reuse-motivation-reuse}), and average
successful-step cost rises in lock-step
(Figure~\ref{fig:reuse-motivation-steps}).  This motivates
augmenting retrieval with a layout-conditioned signal that uses
the current observation as additional evidence for reuse beyond
semantic similarity.

\section{Design}
\label{sec:design}

\SkillMigrator{} reduces
$\mathcal{N}(s_i,\tilde{o}_{i-1}\mid\mathcal{K},\pi_\theta)$ in the
objective of \S\ref{sec:background} by reusing skills in $\mathcal{K}$
across websites and domains whose surface wording and labels differ.
At each pair $(s_i,\tilde{o}_{i-1})$, the agent (i) summarises the
live snapshot and retrieves the most similar skill from $\mathcal{K}$
(\S\ref{sec:design:representation}); (ii) parses one \emph{value} for
each slot of that skill from the user's instruction; 
and (iii) binds each slot to a concrete
control on the live page so that the skill expands into grounded
primitive actions without an LLM call (\S\ref{sec:design:value}). Figure~\ref{fig:architecture} summarises this runtime control flow. 

\subsection{Skill Record and Retrieval}
\label{sec:design:representation}

\paragraph{Snapshot.}
At time $t$, the observation $o_t$ is a Playwright accessibility
snapshot, a YAML-like serialization of the agent-visible accessibility
tree~\citep{drouin2024workarena}.  Each node exposes a semantic role
(\texttt{textbox}, \texttt{button}, \texttt{link}, \texttt{heading},
\dots), an accessible name, and a stable agent-addressable reference
\texttt{ref=eN} that primitive actions use as their target.  A
deterministic rule scans $o_t$ once and produces (i) a one-line
\emph{page summary} $\rho(o_t)$ that lists the page heading, the
labelled controls inside the principal form, and the primary button
labels, and (ii) the list $V(o_t)$ of all interactive nodes with their
roles and names.

\paragraph{Skill record.}
Rather than storing a literal action recipe, which would not survive
relabelling on a new site, each induced skill is stored as a transferable interaction pattern (\TIP{})---a tuple
\begin{equation}
  k \;=\; \bigl(\,\iota_k,\; \sigma_k,\; \Phi_k,\; \tau_k\,\bigr),
  \label{eq:skill}
\end{equation}
where $\iota_k$ is a one-sentence natural-language \emph{intent},
$\sigma_k$ is the skill's \emph{operation template}, drawn from a
set $\Sigma$ of templates mined offline by clustering training
trajectories on action shape (e.g.\ a single \texttt{fill} on a
\texttt{searchbox} followed by a submit, or several \texttt{fill}s
followed by a labelled \texttt{button}),
$\Phi_k=\{\xi_1,\dots,\xi_m\}$ is the \emph{slot schema}, and
$\tau_k$ is the cleaned accessibility-tree skeleton of the
induction-time snapshot, kept as a small labelled tree carrying
role and name per node and used by the layout signal in
Eq.~\eqref{eq:score}.  Each slot $\xi$ stores a key (e.g.\
\emph{post\_title}), a one-line descriptor $d_\xi$ such as
\emph{``the headline of the new entry''}, and a small synonym set
$T_\xi$ mined from co-clustered training trajectories
(e.g.\ \emph{title}~$\to$~\{\emph{title, headline, subject, name,
summary, topic}\}).  The record contains no \texttt{ref=eN}
identifiers, which are properties of the live page at test time.  Each
operation template $\sigma$ comes with a fixed deterministic
\emph{plan} that calls \texttt{fill}/\texttt{click}/\texttt{press}
primitives once $\Phi_k$ is bound (Algorithm~\ref{alg:inference}).
Reusing the same template across sites is what makes a skill
cross-domain rather than a per-trajectory replay.

%
%
%
%

\definecolor{srcBlue}{HTML}{2E5AAC}
\definecolor{tgtRed}{HTML}{B53A3A}
\definecolor{skillOrange}{HTML}{D97706}
\definecolor{srcBg}{HTML}{F1F5FC}
\definecolor{srcBorder}{HTML}{B6C4DC}
\definecolor{tgtBg}{HTML}{FCF2F2}
\definecolor{tgtBorder}{HTML}{E0B6B6}
\definecolor{skillBg}{HTML}{FFF7E6}
\definecolor{skillBorder}{HTML}{D97706}
\definecolor{instrBg}{HTML}{FAF4FB}
\definecolor{instrBorder}{HTML}{C9A6CF}
\definecolor{slotTitle}{HTML}{1F77B4}
\definecolor{slotBody}{HTML}{2CA02C}
\definecolor{slotSubmit}{HTML}{D62728}

\providecommand{\boxTitle}{\bfseries\scriptsize}

\providecommand{\fdColWside}{4.05}
\providecommand{\fdColWmid}{4.55}
\providecommand{\fdGapMid}{0.85}
\providecommand{\fdTopH}{5.20}
\providecommand{\fdBotGap}{0.85}
\providecommand{\fdImgW}{3.65}

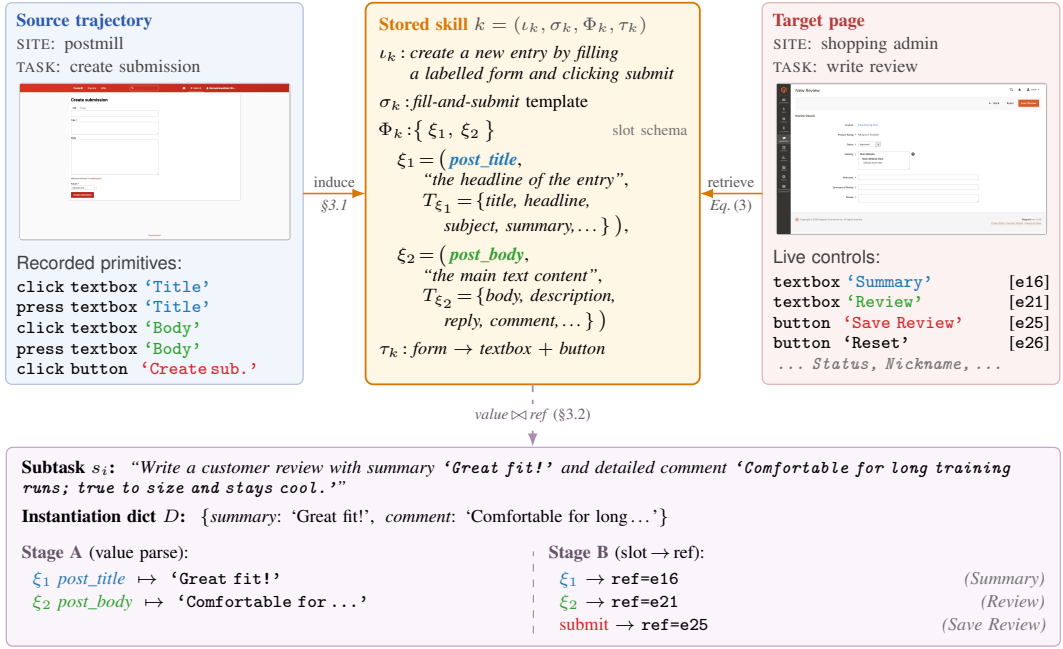
\begin{figure}[t]
  \centering
  \resizebox{\linewidth}{!}{%
  \begin{tikzpicture}[
      font=\scriptsize,
      every node/.style={align=left, inner sep=0pt},
      sectLab/.style={font=\tiny\itshape, text=black!65, inner sep=1pt},
      arrInd/.style={-{Latex[length=1.8mm]}, line width=0.6pt,
                     draw=skillOrange},
      arrBind/.style={-{Latex[length=1.8mm]}, line width=0.7pt,
                      draw=instrBorder!85!black, dashed},
      imgFrame/.style={draw=black!30, line width=0.3pt,
                       rounded corners=1pt, inner sep=0.4pt},
  ]

    \coordinate (srcTL) at (0,0);
    \coordinate (srcBR) at (\fdColWside, -\fdTopH);
    \coordinate (skTL)  at (\fdColWside + \fdGapMid, 0);
    \coordinate (skBR)  at (\fdColWside + \fdGapMid + \fdColWmid, -\fdTopH);
    \coordinate (tgtTL) at (\fdColWside + \fdGapMid + \fdColWmid + \fdGapMid, 0);
    \coordinate (tgtBR) at (\fdColWside + 2*\fdGapMid + \fdColWmid + \fdColWside,
                            -\fdTopH);

    \draw[srcBorder, fill=srcBg, line width=0.5pt, rounded corners=2.5pt]
      (srcTL) rectangle (srcBR);
    \node[anchor=north west, text width=\fdColWside cm-8pt,
          align=left, font=\scriptsize, inner sep=4pt] at (srcTL) {%
      {\color{srcBlue}\boxTitle Source trajectory}\\[1pt]
      {\color{black!72}\scriptsize\textsc{site}: postmill\,\\[1pt]
       \textsc{task}: create submission}\\[3pt]
      \begin{minipage}{\linewidth}\centering
        \tikz\node[imgFrame]{\includegraphics[width=\fdImgW cm]%
          {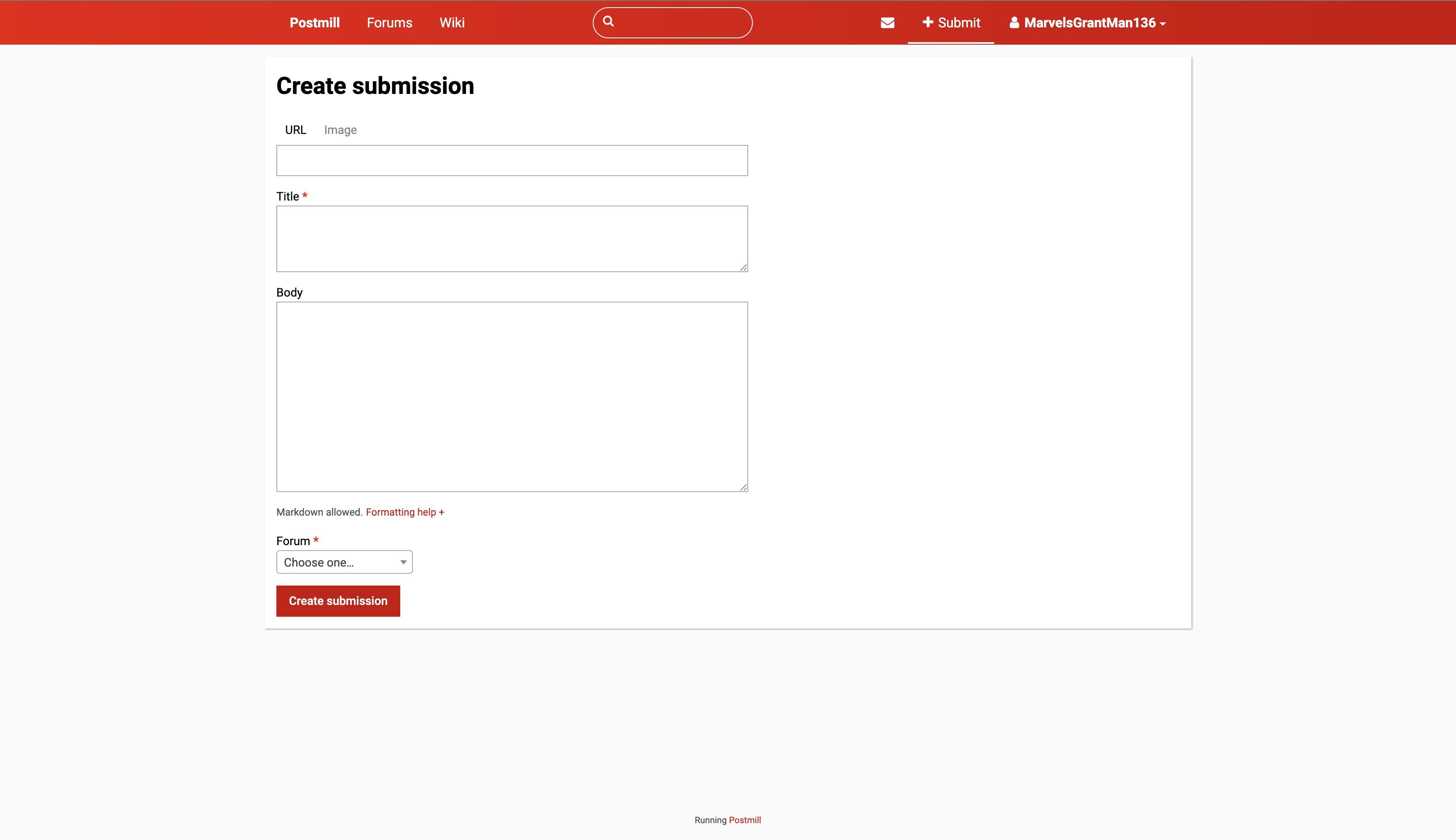}};
      \end{minipage}\\[5pt]
      {\sffamily\scriptsize\color{black!75}Recorded primitives:}\\[1pt]
      \texttt{\scriptsize click  textbox \textcolor{slotTitle}{`Title'}}\\
      \texttt{\scriptsize press  textbox \textcolor{slotTitle}{`Title'}}\\
      \texttt{\scriptsize click  textbox \textcolor{slotBody}{`Body'}}\\
      \texttt{\scriptsize press  textbox \textcolor{slotBody}{`Body'}}\\
      \texttt{\scriptsize click  button~ \textcolor{slotSubmit}{`Create\,sub.'}}%
    };

    \draw[skillBorder, fill=skillBg, line width=0.7pt, rounded corners=3pt]
      (skTL) rectangle (skBR);
    \node[anchor=north west, text width=\fdColWmid cm-10pt,
          align=left, font=\scriptsize, inner sep=5pt,
          execute at begin node={\hyphenpenalty=10000\exhyphenpenalty=10000}]
          at (skTL) {%
      {\color{skillOrange}\boxTitle Stored skill}~%
      {\color{black!70}$k=(\iota_k,\sigma_k,\Phi_k,\tau_k)$}\\[3pt]
      \makebox[1.55em][l]{$\iota_k\!:$}\,\emph{create a new entry by filling}\\
      \makebox[1.55em][l]{}\,\emph{a labelled form and clicking submit}\\[3pt]
      \makebox[1.55em][l]{$\sigma_k\!:$}\,\,\emph{fill-and-submit} template\\[3pt]
      \makebox[1.55em][l]{$\Phi_k\!:$}\,\,$\bigl\{\,\xi_1,\,\xi_2\,\bigr\}$%
      \hfill{\color{black!55}\tiny slot schema}\\[2pt]
      \hspace*{1.0em}$\xi_1\!=\!\bigl(\,$%
        \textcolor{slotTitle}{\textbf{\emph{post\_title}}},\\
      \hspace*{2.4em}\emph{``the headline of the entry''},\\
      \hspace*{2.4em}$T_{\xi_1}\!=\!\{$\emph{title, headline,}\\
      \hspace*{3.6em}\emph{subject, summary,\,\dots}$\}\,\bigr)$,\\[2pt]
      \hspace*{1.0em}$\xi_2\!=\!\bigl(\,$%
        \textcolor{slotBody}{\textbf{\emph{post\_body}}},\\
      \hspace*{2.4em}\emph{``the main text content''},\\
      \hspace*{2.4em}$T_{\xi_2}\!=\!\{$\emph{body, description,}\\
      \hspace*{3.6em}\emph{reply, comment,\,\dots}$\}\,\bigr)$\\[3pt]
      \makebox[1.55em][l]{$\tau_k\!:$}%
        \,\,\mbox{\emph{form} $\to$ \emph{textbox} $+$ \emph{button}}%
    };

    \draw[tgtBorder, fill=tgtBg, line width=0.5pt, rounded corners=2.5pt]
      (tgtTL) rectangle (tgtBR);
    \node[anchor=north west, text width=\fdColWside cm-8pt,
          align=left, font=\scriptsize, inner sep=4pt] at (tgtTL) {%
      {\color{tgtRed}\boxTitle Target page}\\[1pt]
      {\color{black!72}\scriptsize\textsc{site}: shopping admin\,\\[1pt]
       \textsc{task}: write review}\\[3pt]
      \begin{minipage}{\linewidth}\centering
        \tikz\node[imgFrame]{\includegraphics[width=\fdImgW cm]%
          {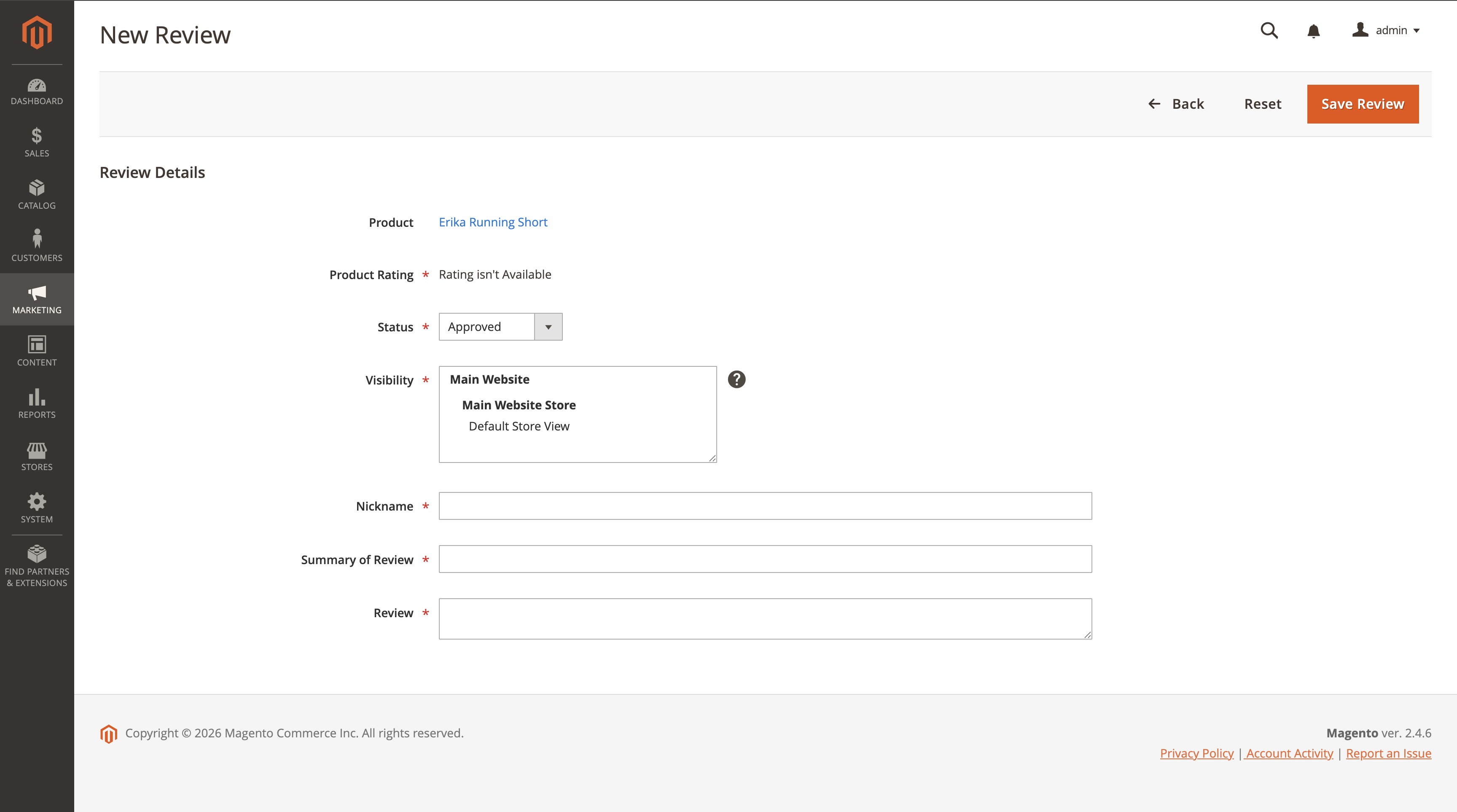}};
      \end{minipage}\\[5pt]
      {\sffamily\scriptsize\color{black!75}Live controls:}\\[1pt]
      \texttt{\scriptsize textbox \textcolor{slotTitle}{`Summary'}%
        \hfill\textnormal{\,[\texttt{e16}]}}\\
      \texttt{\scriptsize textbox \textcolor{slotBody}{`Review'}~%
        \hfill\textnormal{\,[\texttt{e21}]}}\\
      \texttt{\scriptsize button~ \textcolor{slotSubmit}{`Save Review'}%
        \hfill\textnormal{\,[\texttt{e25}]}}\\
      \texttt{\scriptsize button~ `Reset'\hfill\textnormal{\,[\texttt{e26}]}}\\
      \texttt{\scriptsize \textcolor{black!55}{\emph{\,\dots\ Status,
        Nickname,\,\dots}}}%
    };

    \pgfmathsetmacro{\arrY}{-\fdTopH/2}
    \coordinate (srcArrR) at (\fdColWside, \arrY);
    \coordinate (skArrL)  at (\fdColWside + \fdGapMid, \arrY);
    \coordinate (skArrR)  at (\fdColWside + \fdGapMid + \fdColWmid, \arrY);
    \coordinate (tgtArrL) at (\fdColWside + 2*\fdGapMid + \fdColWmid, \arrY);
    \draw[arrInd] (srcArrR) -- (skArrL)
          node[sectLab, midway, above=1pt]{\emph{induce}}
          node[sectLab, midway, below=1pt]{\S\ref{sec:design:representation}};
    \draw[arrInd] (tgtArrL) -- (skArrR)
          node[sectLab, midway, above=1pt]{\emph{retrieve}}
          node[sectLab, midway, below=1pt]{Eq.\,\eqref{eq:score}};

    \pgfmathsetmacro{\botY}{-\fdTopH - \fdBotGap}
    \coordinate (botTL) at (0, \botY);
    \pgfmathsetmacro{\botW}{\fdColWside + 2*\fdGapMid + \fdColWmid + \fdColWside}

    \node[anchor=north west,
          draw=instrBorder, fill=instrBg, line width=0.5pt,
          rounded corners=2.5pt,
          text width=\botW cm-12pt,
          align=left, font=\scriptsize,
          inner xsep=6pt, inner ysep=5pt] at (botTL) (ins) {%
      \textbf{Subtask $s_i$:}\;
      \emph{``Write a customer review with summary
        \texttt{`Great~fit!'} and detailed comment
        \texttt{`Comfortable for long training runs;
                true to size and stays cool.'}''}\\[3pt]
      \textbf{Instantiation dict $D$:}\;
      $\{$\emph{summary}: `Great fit!',\;
        \emph{comment}: `Comfortable for long\,\dots'$\}$\\[6pt]
      \begin{minipage}[t]{0.485\linewidth}
        \textbf{\color{instrBorder!50!black}Stage A}~%
          \textnormal{(value parse):}\\[2pt]
        \hspace*{0.6em}\textcolor{slotTitle}{$\xi_1$ \emph{post\_title}}
          $\;\mapsto\;$ \texttt{`Great fit!'}\\[1pt]
        \hspace*{0.6em}\textcolor{slotBody}{$\xi_2$ \emph{post\_body}}
          $\;\mapsto\;$ \texttt{`Comfortable for\,\dots'}
      \end{minipage}%
      \hfill\hfill%
      \begin{minipage}[t]{0.485\linewidth}
        \textbf{\color{instrBorder!50!black}Stage B}~%
          \textnormal{(slot\,$\to$\,ref):}\\[2pt]
        \hspace*{0.6em}\textcolor{slotTitle}{$\xi_1$}
          $\to$ \texttt{ref=e16}\hfill{\color{black!55}\emph{(Summary)}}\\[1pt]
        \hspace*{0.6em}\textcolor{slotBody}{$\xi_2$}
          $\to$ \texttt{ref=e21}\hfill{\color{black!55}\emph{(Review)}}\\[1pt]
        \hspace*{0.6em}\textcolor{slotSubmit}{submit}
          $\to$ \texttt{ref=e25}\hfill{\color{black!55}\emph{(Save Review)}}
      \end{minipage}%
    };

    \draw[dashed, line width=0.3pt, draw=instrBorder!75!black]
      ($(ins.south) + (0, 0.18)$) -- ($(ins.south) + (0, 1.30)$);

    \pgfmathsetmacro{\skXmid}{\fdColWside + \fdGapMid + \fdColWmid/2}
    \draw[arrBind] (\skXmid, -\fdTopH) -- (\skXmid, \botY)
         node[midway, font=\tiny, fill=white, inner sep=2pt,
              text=instrBorder!55!black]
              {\emph{value\,$\bowtie$\,ref}\;(\S\ref{sec:design:value})};

  \end{tikzpicture}%
  }

  \caption{End-to-end cross-domain example.  A skill induced from a
Postmill \emph{create-submission} trajectory (left) is retrieved on a
shopping-admin \emph{write-review} page (right): \emph{post\_title} and
\emph{post\_body} rebind to \emph{Summary} and \emph{Review} through
the synonym pools in $\Phi_k$, and the bottom panel walks through
Stage~A and Stage~B for one subtask.  Same-colour fields
(\textcolor{slotTitle}{\emph{title}-like},
\textcolor{slotBody}{\emph{body}-like},
\textcolor{slotSubmit}{\emph{submit}}) denote the same abstract slot
across both pages.}
  \label{fig:design-example}
\end{figure}

\paragraph{Retrieval.}
Given the current pair $(s_i,\tilde{o}_{i-1})$, we score each skill
$k\in\mathcal{K}$ by combining a \emph{text} signal that captures
\emph{what kind of subtask} is being attempted with a
\emph{layout} signal that captures \emph{what kind of page} the
agent is on:
\begin{equation}
  \mathrm{score}(k,\,s_i,\,\tilde{o}_{i-1})
  \;=\;
  \alpha\,
  \underbrace{
    \mathbf{e}\bigl(s_i \,\Vert\, \rho(\tilde{o}_{i-1})\bigr)^{\!\top}
    \mathbf{e}(\delta_k)
  }_{\text{text signal}}
  \;+\;
  (1-\alpha)\,
  \underbrace{
    \mathcal{L}\bigl(k,\,\tilde{o}_{i-1}\bigr)
  }_{\text{layout signal}},
  \label{eq:score}
\end{equation}
with $\alpha\!\in\![0,1]$.  Here $\mathbf{e}(\cdot)$ is a frozen
sentence encoder~\citep{reimers2019sentence}, and
$\delta_k=\iota_k\,\Vert\,\bigcup_{\xi\in\Phi_k}(d_\xi\,\Vert\,T_\xi)$
is a \emph{rich descriptor} concatenating the skill's intent with
every slot descriptor and synonym, so paraphrased subtasks across
sites stay close in embedding space~\citep{liu2020coach,wang2021bridge}.
The layout signal $\mathcal{L}(k,\tilde{o}_{i-1})$ uses the live
accessibility-tree structure of $\tilde{o}_{i-1}$ to ground retrieval
in what the page actually exposes. The \emph{tree edit distance (TED)} term
$1\!-\!\mathrm{TED}(\tau_k,\tau(\tilde{o}_{i-1}))/\!\max(|\tau_k|,
|\tau(\tilde{o}_{i-1})|)$ is
computed by APTED on small trees~\citep{tai1979tree,pawlik2016tree}. This is the dominant signal when source and target sites use
different labels for structurally analogous forms.




We open \emph{skill mode} only if
$\mathrm{score}(k^\star,s_i,\tilde{o}_{i-1})\!\geq\!\beta$ for the
top-1 skill $k^\star$; otherwise the agent stays in react
mode~\citep{yao2023react}.  This gate matters because not every test
subtask has a transferable analog in $\mathcal{K}$, and forcing a
weakly matched skill on an unrelated page would silently corrupt the
trajectory.  $\alpha$ and $\beta$ are tuned on held-out training
trajectories. 

\subsection{Slot Binding and Execution}
\label{sec:design:value}

Once $k^\star$ is chosen, the agent must associate each slot
$\xi\in\Phi_{k^\star}$ with a concrete \emph{value} string before
binding it to a control on the page.  We follow the cross-domain
slot-filling view of \citet{liu2020coach,wang2021bridge}: a slot is
identified by its NL descriptor $d_\xi$ and synonym pool $T_\xi$, not
by its training-time key, so unseen vocabulary on the test side does
not break the match.

\begin{algorithm}[!t]
  \caption{\SkillMigrator{} inference for one subtask
  $(s_i,\tilde{o}_{i-1})$.}
  \label{alg:inference}
  \begin{algorithmic}[1]
    \Require subtask $s_i$, snapshot $\tilde{o}_{i-1}$, library
             $\mathcal{K}$, encoder $\mathbf{e}$, instantiation dict
             $D$, gate threshold $\beta$.
    \State compute summary $\rho(\tilde{o}_{i-1})$ and node list
           $V(\tilde{o}_{i-1})$
    \State $k^\star \gets \arg\max_{k\in\mathcal{K}}
            \mathrm{score}(k,s_i,\tilde{o}_{i-1})$ via Eq.~\eqref{eq:score}
    \If{$\mathrm{score}(k^\star,s_i,\tilde{o}_{i-1})<\beta$}
        \State \Return $\pi_\theta(s_i,\tilde{o}_{i-1})$
        \Comment{fall back to react mode}
    \EndIf
    \State \emph{Stage A}: Hungarian-solve $\Phi_{k^\star}\!\times\! D$
           for slot values; for any unbound slot, extract a span from $s_i$
    \State \emph{Stage B}: Hungarian-solve
           $\Phi_{k^\star}\!\times\! V(\tilde{o}_{i-1})$ for slot$\to$ref bindings
    \State emit primitives per the plan of $\sigma_{k^\star}$;
           escalate any unbound \emph{required} slot to $\pi_\theta$
  \end{algorithmic}
\end{algorithm}

\definecolor{srcBlue}{HTML}{2E5AAC}
\definecolor{tgtRed}{HTML}{B53A3A}
\definecolor{skillOrange}{HTML}{D97706}
\definecolor{srcBg}{HTML}{F1F5FC}
\definecolor{srcBorder}{HTML}{B6C4DC}
\definecolor{skillBg}{HTML}{FFF7E6}
\definecolor{skillBorder}{HTML}{D97706}
\definecolor{instrBg}{HTML}{FAF4FB}
\definecolor{instrBorder}{HTML}{C9A6CF}
\definecolor{tgtBg}{HTML}{FCF2F2}
\definecolor{tgtBorder}{HTML}{E0B6B6}
\definecolor{gateBg}{HTML}{FFF3D6}
\definecolor{gateBorder}{HTML}{C9941A}
\definecolor{reactBg}{HTML}{F0F0F0}
\definecolor{reactBorder}{HTML}{A8A8A8}
\definecolor{passColor}{HTML}{2D7A4F}
\definecolor{failColor}{HTML}{B53A3A}
\definecolor{loopColor}{HTML}{6F6F6F}

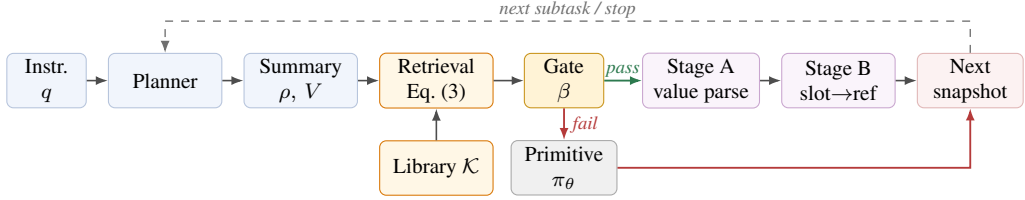
\begin{figure}[t]
  \centering
  \begin{tikzpicture}[
      font=\footnotesize,
      box/.style={draw, rounded corners=2.5pt, align=center,
                  inner xsep=4pt, inner ysep=2pt,
                  minimum height=0.72cm,
                  minimum width=1.50cm,
                  line width=0.4pt},
      arrMain/.style={-{Latex[length=1.6mm]}, line width=0.6pt,
                      draw=black!70},
      arrPass/.style={-{Latex[length=1.6mm]}, line width=0.7pt,
                      draw=passColor},
      arrFail/.style={-{Latex[length=1.6mm]}, line width=0.7pt,
                      draw=failColor},
      arrLoop/.style={-{Latex[length=1.6mm]}, line width=0.5pt,
                      draw=loopColor, dashed},
      pathLab/.style={font=\scriptsize\itshape, inner sep=1pt},
  ]

    \node[box, fill=srcBg,    draw=srcBorder,
          minimum width=1.05cm]  (q)   {Instr.\\$q$};
    \node[box, fill=srcBg,    draw=srcBorder, right=0.28cm of q]    (pl)  {Planner};
    \node[box, fill=srcBg,    draw=srcBorder, right=0.28cm of pl]   (sum) {Summary\\$\rho,\,V$};
    \node[box, fill=skillBg,  draw=skillBorder, right=0.28cm of sum] (ret){Retrieval\\Eq.~\eqref{eq:score}};
    \node[box, fill=gateBg,   draw=gateBorder, right=0.40cm of ret,
          minimum width=1.05cm] (gate) {Gate\\$\beta$};
    \node[box, fill=instrBg,  draw=instrBorder, right=0.50cm of gate] (val) {Stage~A\\value parse};
    \node[box, fill=instrBg,  draw=instrBorder, right=0.28cm of val] (bind){Stage~B\\slot$\to$ref};
    \node[box, fill=tgtBg,    draw=tgtBorder, right=0.28cm of bind,
          minimum width=1.40cm] (env) {Next\\snapshot};

    \node[box, fill=skillBg, draw=skillBorder, below=0.42cm of ret]
          (lib) {Library $\mathcal{K}$};
    \node[box, fill=reactBg, draw=reactBorder, below=0.42cm of gate,
          minimum width=1.40cm]
          (react) {Primitive\\$\pi_\theta$};

    \draw[arrMain] (q)   -- (pl);
    \draw[arrMain] (pl)  -- (sum);
    \draw[arrMain] (sum) -- (ret);
    \draw[arrMain] (ret) -- (gate);
    \draw[arrMain] (lib.north) -- (lib.north |- ret.south);

    \draw[arrPass] (gate) -- node[pathLab, above, text=passColor]{pass}
                              (val);
    \draw[arrMain] (val)  -- (bind);
    \draw[arrMain] (bind) -- (env);

    \draw[arrFail] (gate) -- node[pathLab, right=2pt, text=failColor]{fail}
                              (react);
    \draw[arrFail] (react.east) -| (env.south);

    \pgfmathsetmacro{\loopY}{0.42}
    \coordinate (envUp) at ($(env.north) + (0,\loopY)$);
    \coordinate (plUp)  at ($(pl.north)  + (0,\loopY)$);
    \draw[arrLoop] (env.north) -- (envUp) -- (plUp)
         node[pathLab, midway, above, text=loopColor]
              {next subtask / stop}
         -- (pl.north);

  \end{tikzpicture}

  \caption{Runtime control flow.  Passing the gate $\beta$ routes the
  subtask through skill mode (Stage~A then Stage~B). Failing falls back
  to a single primitive step from $\pi_\theta$.  
  }
  \label{fig:architecture}
\end{figure}

\paragraph{Stage A: parsing slot values from the instruction.} The planner forces each subtask $s_i$ to expose an \emph{instantiation dict}
$D=\{(\eta_j,v_j)\}_j$ in the task spec~\citep{zhou2024webarena},
where $\eta_j$ is a key (\emph{subject}, \emph{message},
\emph{from\_location}, \dots) and $v_j$ is the corresponding value.
We score every candidate pair $(\xi,\eta_j)$ by
\begin{equation*}
  \mathbf{e}(d_\xi\,\Vert\, T_\xi)^{\!\top}
  \mathbf{e}(\eta_j)
  \;+\;
  \lambda_{\mathrm{lit}}\!\cdot\!\mathbb{1}[\eta_j\in T_\xi],
\end{equation*}
and run a Hungarian
solve~\citep{kuhn1955hungarian,munkres1957algorithms} on the resulting
$|\Phi_{k^\star}|\!\times\!|D|$ matrix to obtain a one-to-one
assignment.  The literal bonus $\lambda_{\mathrm{lit}}$ is the same
disambiguation mechanism long used in slot
filling~\citep{bapna2017towards}: when $\eta_j$ already appears
verbatim inside $T_\xi$ the match is essentially certain.  In the
example of Figure~\ref{fig:design-example}, \emph{summary} appears
in $T_{\mathit{post\_title}}$ and \emph{comment} in
$T_{\mathit{post\_body}}$, so the source-trajectory keys
(\emph{post\_title}, \emph{post\_body}) are correctly bound to the
target keys (\emph{summary}, \emph{comment}).

When $D$ is empty or covers fewer slots than $\Phi_{k^\star}$ needs,
we extract candidate spans $\mathcal{C}(s_i)$ directly from the
instruction text using a fixed cue grammar (quoted strings, list
literals, dates, URLs, \emph{from~X~to~Y}, \emph{titled~X},
\emph{named~X}, capitalised proper-noun phrases, \dots).  Each span
$c$ comes with a small prior reflecting how distinctive its cue is
(e.g.\ a quoted string is a stronger signal than a bare
capitalisation).  We then score
$\mathbf{e}(d_\xi\,\Vert\, T_\xi)^{\!\top}\mathbf{e}(\mathrm{ctx}(c))$
on the surrounding context window with the candidate replaced by a
placeholder, and run a Hungarian assignment.  The two paths are run
in order: dict-match first, then instruction-extract for any
still-unbound slot.

\paragraph{Stage B: binding each slot to a control on the page.} The bound values must finally be typed into actual page controls.
For every interactive node $v\in V(\tilde{o}_{i-1})$ we build a
deterministic \emph{control descriptor}
$d(v)=\mathrm{role}(v)\,\Vert\,\mathrm{name}(v)$, score every
(slot, control) pair by
$\mathbf{e}(d_\xi\,\Vert\,T_\xi)^{\!\top}\mathbf{e}(d(v))$, restrict
to controls of compatible role (textbox-like for fillable slots), and
run a second Hungarian solve.  An assignment is accepted only above a
similarity threshold. Below it we declare the slot \emph{unbound} and
fall back to $\pi_\theta$ for that slot.

The critical effect of this stage is that the \emph{number of fills}
emitted on the live page is determined by the live page and the
matched slots, not by the action count of the source trajectory.  On
a five-field form like \emph{Create new forum} (Name, Title,
Description, Sidebar, Tags), an instantiation dict that supplies only
three values fills only the three matching textboxes and leaves the
other two empty; on a two-field form like \emph{New issue} (Title,
Description) the same skill emits two fills.  This handles the
variable-arity issue without touching the stored skill record.


\paragraph{Plan execution.}
Each operation template $\sigma\in\Sigma$ has a fixed plan keyed on
the bound slots.  A \emph{search} template emits a
\texttt{fill} on the search input and a \texttt{click} on the submit button,
or a \texttt{press(Enter)} when no submit button binds.  A
\emph{fill-and-submit} template emits one \texttt{fill} per bound
slot in declared order, then a \texttt{click} on the button whose
label matches the global submit-keyword pool (\emph{create, submit,
post, send, save, update, publish}) mined from training.  A
\emph{click-by-text} template emits a single \texttt{click} on the
link or cell whose name best matches the target value.  
Any unbound \emph{required} slot escalates to $\pi_\theta$.


\section{Experiments}
\label{sec:experiments}
We evaluate \SkillMigrator{} with a focus on the LLM-action count
$\mathcal{N}(s_i,\tilde{o}_{i-1}\mid\mathcal{K},\pi_\theta)$ from
Eq.~(1) of \S\ref{sec:background}, corresponding to the
\emph{average successful tool steps} metric used in prior
work~\citep{wang2025inducing,yu2026polyskill}.  Three research questions
organise the evaluation:
\begin{itemize}[topsep=0pt,itemsep=0pt,parsep=0pt,partopsep=0pt,leftmargin=*]
  \item \textbf{RQ1.} Does \SkillMigrator{} reduce $\mathcal{N}$ while preserving task success rate on different splits where previous baselines are restricted to within-domain reuse?
  \item \textbf{RQ2.} Does \SkillMigrator{}
  \emph{compose} with existing skill libraries: when its gate
  falls back, does plugging an ASI or PolySkill library underneath
  add to the gain?
  \item \textbf{RQ3.} Which component is responsible for the gain, and how sensitive are the results
  to the weight $\alpha$ and gate threshold $\beta$?
\end{itemize}

\subsection{Setup}
\label{sec:exp:setup}
\paragraph{Benchmarks and baselines.}
We evaluate on \emph{Mind2Web}~\citep{deng2023mind2web} (137
websites, 31 domains, cross-task / cross-website / cross-domain
splits) and \emph{WebArena}~\citep{zhou2024webarena} (812
executable tasks across 6 columns).  Baselines are 
\emph{ReAct}~\citep{yao2023react} (no skill library),
SkillWeaver~\citep{zheng2025skillweaver},
ASI~\citep{wang2025inducing}, and
PolySkill~\citep{yu2026polyskill}.  ASI, PolySkill, and
\SkillMigrator{} are each reported in both a \emph{static} regime
(library fixed before evaluation) and an \emph{+Update} regime
(skills induced online during evaluation), following PolySkill. 
 
\paragraph{Metrics.}
The main metric is the \emph{average LLM-action count} on
successful trajectories,
$\bar{\mathcal{N}}=\mathbb{E}_{q\in\mathcal{D}_{\mathrm{succ}}}
\sum_i\mathcal{N}(s_i,\tilde{o}_{i-1}\mid\mathcal{K},\pi_\theta)$,
counting one LLM call per primitive tool action and zero per
retrieved skill. We additionally
report task success rate (SR), skill reuse rate, and library size $|\mathcal{K}|$.
 
\paragraph{Configuration.}
Unless stated otherwise we use $\alpha=0.6$ for the text/layout
mixing weight in Eq.~(3) and $\beta=0.20$ for the gate threshold
of \S\ref{sec:design:representation}.  Both were selected on a 10\% subset of training trajectories without overlapping with any
test set.  More details are in Appendix~\ref{app:implementation}.

\subsection{Main Results}
\label{sec:exp:main}
\begin{figure}[t]
  \centering
  \begin{subfigure}[b]{0.48\linewidth}
    \centering
    \includegraphics[width=\linewidth]{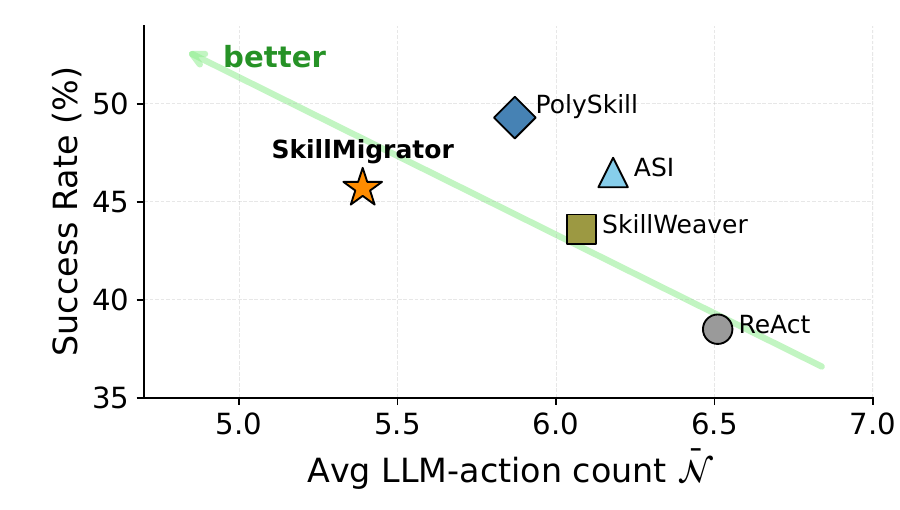}
    \caption{WebArena average.}
    \label{fig:pareto-wa}
  \end{subfigure}\hfill
  \begin{subfigure}[b]{0.48\linewidth}
    \centering
    \includegraphics[width=\linewidth]{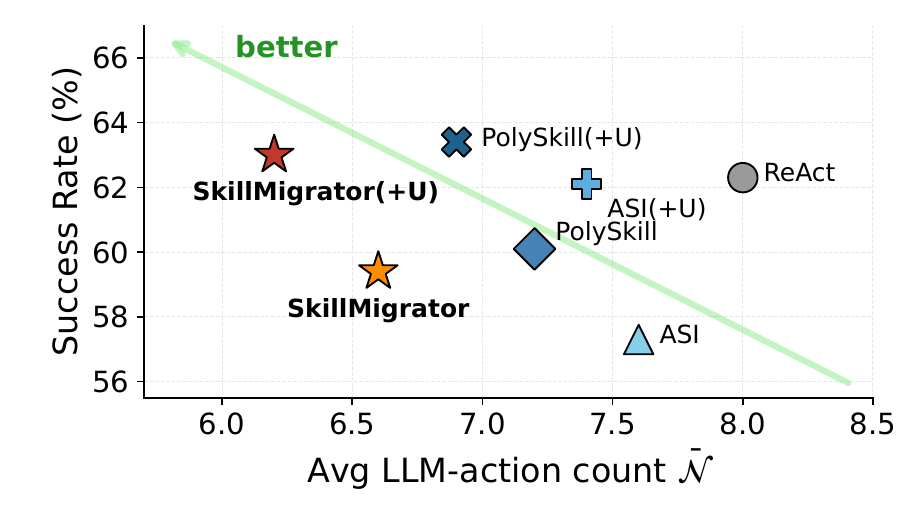}
    \caption{Mind2Web cross-domain.}
    \label{fig:pareto-m2w}
  \end{subfigure}
  \caption{Success rate against average LLM-action count
  $\bar{\mathcal{N}}$ (lower-left is better on both axes).
  \SkillMigrator{} sits left of every baseline at comparable
  SR on both benchmarks, using fewer LLM calls per
  successful trajectory.}
  \label{fig:sr-vs-N}
\end{figure}

\begin{table*}[t]
  \centering
  \footnotesize
  \caption{WebArena per-domain task success rate (SR, \%) and
  average LLM-action count $\bar{\mathcal{N}}$ on successful
  trajectories. $\bar{\mathcal{N}}$ rows are
  extracted from trajectory logs under one unified counter.}
  \label{tab:webarena}
  \resizebox{\textwidth}{!}{%
  \begin{tabular}{lcc cc cc cc cc cc cc}
    \toprule
     & \multicolumn{2}{c}{\textsc{Shop}}
     & \multicolumn{2}{c}{\textsc{Admin}}
     & \multicolumn{2}{c}{\textsc{Reddit}}
     & \multicolumn{2}{c}{\textsc{GitLab}}
     & \multicolumn{2}{c}{\textsc{Map}}
     & \multicolumn{2}{c}{\textsc{Multi}}
     & \multicolumn{2}{c}{\textsc{Avg.}} \\
    \cmidrule(lr){2-3}\cmidrule(lr){4-5}\cmidrule(lr){6-7}
    \cmidrule(lr){8-9}\cmidrule(lr){10-11}\cmidrule(lr){12-13}
    \cmidrule(lr){14-15}
    \textsc{Method}
     & SR & $\bar{\mathcal{N}}$
     & SR & $\bar{\mathcal{N}}$
     & SR & $\bar{\mathcal{N}}$
     & SR & $\bar{\mathcal{N}}$
     & SR & $\bar{\mathcal{N}}$
     & SR & $\bar{\mathcal{N}}$
     & SR & $\bar{\mathcal{N}}$ \\
    \midrule
    ReAct~\citep{yao2023react}
     & 37.4 & 6.1 & 44.0 & 7.0
     & 66.0 & 5.0 & 38.9 & 7.5
     & 16.4 & 4.5 & 10.3 & 10.5
     & 38.5 & 6.5 \\
    SkillWeaver~\citep{zheng2025skillweaver}
     & 39.3 & 5.7 & 48.2 & 6.5
     & 71.2 & 4.7 & 50.3 & 7.0
     & 17.2 & 4.2 & 16.3 & 9.8
     & 43.6 & 6.1 \\
    ASI~\citep{wang2025inducing}
     & 46.3 & 5.8 & 53.6 & 6.6
     & 73.7 & 4.8 & 46.8 & 7.1
     & 21.5 & 4.3 & 15.1 & 10.0
     & 46.5 & 6.2 \\
    PolySkill~\citep{yu2026polyskill}
     & 51.4 & 5.5 & 54.8 & 6.3
     & 73.2 & 4.5 & 54.2 & 6.8
     & 18.9 & 4.0 & 18.9 & 9.5
     & 49.3 & 5.9 \\
    \midrule
    \SkillMigrator{}
     & 45.5 & \textbf{4.8} & 52.7 & \textbf{5.6}
     & 72.6 & \textbf{4.5} & 46.7 & \textbf{6.7}
     & 19.3 & \textbf{3.0} & 16.7 & \textbf{9.4}
     & 45.7 & \textbf{5.4} \\
    \bottomrule
  \end{tabular}}
\end{table*}
\begin{table*}[t]
  \centering
  \footnotesize
  \caption{Mind2Web success rate, average LLM-action count
  $\bar{\mathcal{N}}$, and library size $|\mathcal{K}|$ on the
  three generalisation splits with GPT-4.1.  Methods are reported
  in two regimes: a \emph{static} library (rows without the
  \emph{(+Update)} suffix) fixed before evaluation, and an
  \emph{+Update} regime (rows tagged \emph{(+Update)}) in which
  new skills are induced online during evaluation. }
  \label{tab:mind2web}
  \resizebox{\textwidth}{!}{%
  \begin{tabular}{l ccc ccc cccc}
    \toprule
     & \multicolumn{3}{c}{\textsc{Cross-task}}
     & \multicolumn{3}{c}{\textsc{Cross-website}}
     & \multicolumn{4}{c}{\textsc{Cross-domain}}
     \\
    \cmidrule(lr){2-4}\cmidrule(lr){5-7}\cmidrule(lr){8-11}
    \textsc{Method}
     & SR & $\bar{\mathcal{N}}$ & $|\mathcal{K}|$
     & SR & $\bar{\mathcal{N}}$ & $|\mathcal{K}|$
     & SR & $\bar{\mathcal{N}}$ & $|\mathcal{K}|$
     & \textsc{Reuse}\,\% \\
    \midrule
    ReAct~\citep{yao2023react}
     & 53.8 & 7.0 & --
     & 56.2 & 7.5 & --
     & 62.3 & 8.0 & --
     & -- \\
    \midrule
    ASI~\citep{wang2025inducing}
     & 52.3 & 6.6 & 50
     & 54.9 & 7.1 & 47
     & 57.3 & 7.6 & 33
     & 17.4 \\
    PolySkill~\citep{yu2026polyskill}
     & 55.4 & 6.3 & 43
     & 57.6 & 6.8 & 44
     & 60.1 & 7.2 & 36
     & 22.7 \\
    \SkillMigrator{}
     & 54.8 & \textbf{5.8} & \textbf{35}
     & 57.1 & \textbf{6.2} & \textbf{38}
     & 59.4 & \textbf{6.6} & \textbf{31}
     & \textbf{28.1} \\
    \midrule
    ASI~\citep{wang2025inducing} (+Update)
     & 59.4 & 6.4 & 66
     & 58.7 & 6.9 & 71
     & 62.1 & 7.4 & 66
     & 23.6 \\
    PolySkill~\citep{yu2026polyskill} (+Update)
     & 63.2 & 6.0 & 47
     & 61.3 & 6.5 & 53
     & 63.4 & 6.9 & 56
     & 31.0 \\
    \SkillMigrator{} (+Update)
     & 62.7 & \textbf{5.5} & \textbf{41}
     & 60.5 & \textbf{5.9} & \textbf{47}
     & 63.0 & \textbf{6.2} & \textbf{49}
     & \textbf{35.4} \\
    \bottomrule
  \end{tabular}}
\end{table*}
Tables~\ref{tab:webarena} and~\ref{tab:mind2web}, together with
the scatter in Figure~\ref{fig:sr-vs-N}, answer RQ1.  The
pattern is consistent across both benchmarks: \SkillMigrator{}
sits at or just below the strongest baseline on SR, but spends
noticeably fewer LLM calls per successful trajectory and reuses
more skills across the test pool.  On WebArena, aggregate SR is
within $-$3.6 points of PolySkill but
$\bar{\mathcal{N}}$ drops from 5.9 to 5.4---an 8.5\% reduction
in policy LLM calls, and 16.9\% against the ReAct baseline of
6.5.  On Mind2Web the same trade-off shows up:
\SkillMigrator{}\,(+Update) achieves a 63.0\% SR on cross-domain at
$\bar{\mathcal{N}}\!=\!6.2$, against PolySkill\,(+Update)'s
63.4\% SR and $\bar{\mathcal{N}}\!=\!6.9$. 
The reuse rate further supports this result: cross-domain reuse rises to
35.4\% versus PolySkill's 31\%, while the library size is consistently
smaller. Overall, \SkillMigrator{} reduces LLM calls per successful
trajectory and improves skill reuse on previously unseen domains without
a significant decrease in accuracy.
\subsection{Orthogonality with Existing Skill Libraries}
\label{sec:exp:orthogonal}
\begin{table}[t]
  \centering
  \footnotesize
  \caption{Orthogonality study (a) and component ablations (b) on the
  WebArena.  
  }
  \label{tab:orth-abl}
  \begin{subtable}[t]{0.45\linewidth}
    \centering
    \caption{WebArena average when the
    gate of \SkillMigrator{} falls back to a secondary library.}
    \label{tab:orthogonal}
    \setlength{\tabcolsep}{4pt}
    \begin{tabular}{lcc}
      \toprule
      \textsc{Configuration} & SR\,\% & $\bar{\mathcal{N}}$ \\
      \midrule
      ASI alone~\citep{wang2025inducing}      & 46.5 & 6.2 \\
      PolySkill alone~\citep{yu2026polyskill} & 49.3 & 5.9 \\
      \SkillMigrator{} alone (gate$\to$ReAct) & 45.7 & 5.4 \\
      \midrule
      \SkillMigrator{} + ASI                  & 47.2 & 5.3 \\
      \SkillMigrator{} + PolySkill            & \textbf{47.7} & \textbf{5.3} \\
      \bottomrule
    \end{tabular}
  \end{subtable}\hfill
  \begin{subtable}[t]{0.52\linewidth}
    \centering
    \caption{Each row removes
    one element of \SkillMigrator{}. Lower $\bar{\mathcal{N}}$ is
    better.}
    \label{tab:ablation}
    \setlength{\tabcolsep}{4pt}
    \begin{tabular}{l cc cc}
      \toprule
       & \multicolumn{2}{c}{\textsc{WA Avg.}}
       & \multicolumn{2}{c}{\textsc{M2W CD}} \\
      \cmidrule(lr){2-3}\cmidrule(lr){4-5}
      \textsc{Variant} & SR\,\% & $\bar{\mathcal{N}}$
                       & SR\,\% & $\bar{\mathcal{N}}$ \\
      \midrule
      Full\,\SkillMigrator{}    & \textbf{45.7} & \textbf{5.4} & \textbf{59.4} & \textbf{6.6} \\
      \,\,text only ($\alpha\!=\!1$)     & 39.2 & 5.7 & 55.8 & 7.3 \\
      \,\,no synonyms                    & 45.4 & 5.4 & 57.0 & 7.1 \\
      \,\,no gate ($\beta\!=\!0$)         & 45.3 & 4.7 & 57.8 & 6.3 \\
      \bottomrule
    \end{tabular}
  \end{subtable}
\end{table}
RQ2 asks whether \SkillMigrator{} \emph{competes} with
baselines or \emph{composes} with them.  When the gate falls
back, the agent does not need to revert to raw ReAct: control can pass to
a secondary library trained by another method.
Table~\ref{tab:orthogonal} shows the result.  The hybrid
\SkillMigrator{}+PolySkill row reaches the lowest
$\bar{\mathcal{N}}$ in the table (5.3, vs.\ 5.4 alone and 5.9
for PolySkill alone), and its 47.7\% SR sits between the two
single-library numbers (45.7 and 49.3).  The two libraries cover different slices of the test set: when a layout
match exists, the layout signal triggers a \SkillMigrator{} skill. When
no such match exists, control routes to PolySkill underneath.  In practice,
\SkillMigrator{} can behave as a retrieval layer that sits on top of
existing skill-induction systems rather than competing with them.

\subsection{Ablation Study}
\label{sec:exp:ablation}
For RQ3, Table~\ref{tab:ablation} isolates three ingredients: the layout
signal $\mathcal{L}$ (\emph{text only} pushes $\alpha$ to 1), the
slot-synonym pool $T_\xi$ (\emph{no synonyms} removes paraphrase
coverage), and the gate (\emph{no gate} sets $\beta\!=\!0$).  The
biggest move is on the WebArena average SR: removing $\mathcal{L}$
drops 6.5 points (45.7$\to$39.2) and lands almost on top of the
no-skill baseline (38.5\%), because instruction text alone cannot
tell apart same-wording, different-structure pages.  On Mind2Web
cross-domain the same ablation costs 3.6 points (59.4$\to$55.8),
and \emph{no synonyms} costs another 2.4 (59.4$\to$57.0):
paraphrase coverage matters most when the target site renames its
labels.  Disabling the gate is the only ablation that lowers
$\bar{\mathcal{N}}$ (5.4$\to$4.7 on WebArena; 6.6$\to$6.3 on
Mind2Web) at almost no SR cost ($-$0.4 on WebArena), so the
default $\beta\!=\!0.20$ errs on the conservative side---a
deployment that tolerates more skill triggers can push for further
LLM-call savings.

\subsection{Sensitivity Analysis}
\label{sec:exp:sensitivity}
\begin{figure}[t]
  \centering
  \begin{subfigure}[b]{0.32\linewidth}
    \centering
    \includegraphics[width=\linewidth]{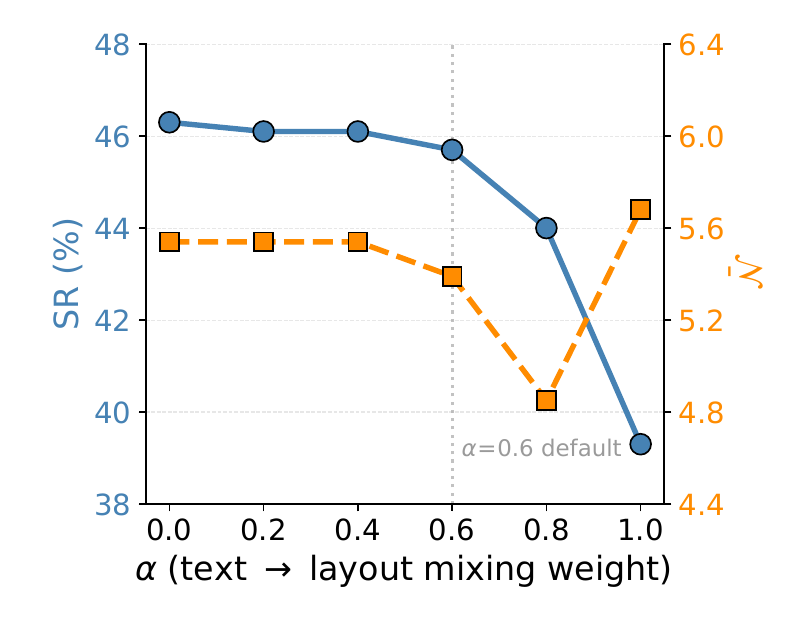}
    \caption{Sweep $\alpha$.}
    \label{fig:sens-alpha}
  \end{subfigure}\hfill
  \begin{subfigure}[b]{0.32\linewidth}
    \centering
    \includegraphics[width=\linewidth]{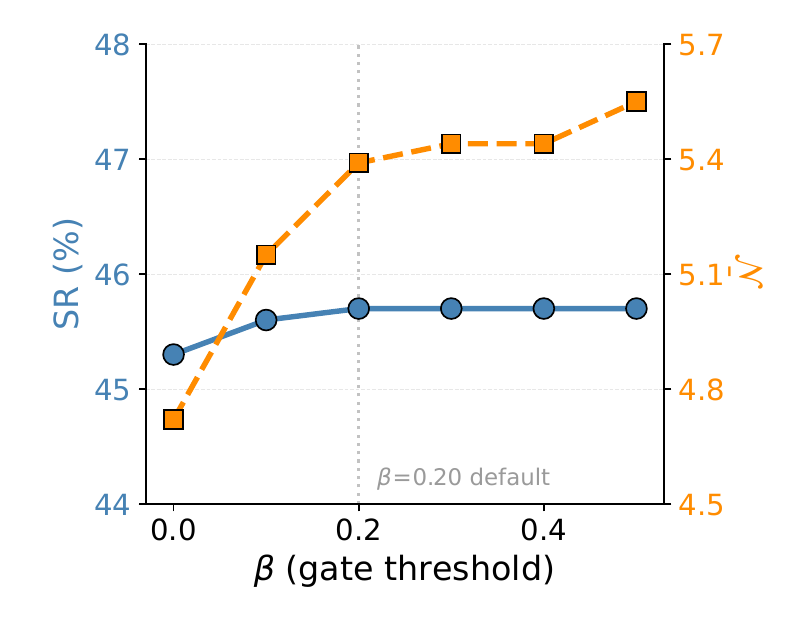}
    \caption{Sweep $\beta$.}
    \label{fig:sens-tau}
  \end{subfigure}\hfill
  \begin{subfigure}[b]{0.32\linewidth}
    \centering
    \includegraphics[width=\linewidth]{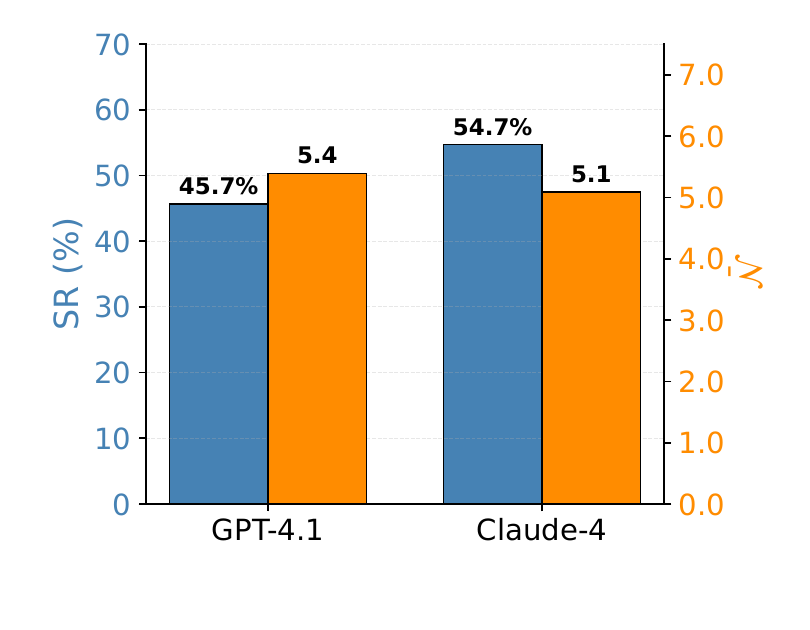}
    \caption{Backbone.}
    \label{fig:sens-backbone}
  \end{subfigure}
  \caption{Sensitivity to the mixing weight $\alpha$, the gate
  threshold $\beta$, and the LLM backbone, on the WebArena. Blue (left axis) is success rate. Orange (right axis)
  is the average LLM-action count $\bar{\mathcal{N}}$.}
  \label{fig:sensitivity}
\end{figure}
We sweep $\alpha\in\{0.0,0.2,\dots,1.0\}$ and
$\beta\in\{0.0,0.1,\dots,0.5\}$ on WebArena with GPT-4.1, and re-run the headline configuration on Claude-4.
Hyperparameters were chosen on a held-out 10\% of training
trajectories, never on the test set.
Figure~\ref{fig:sens-alpha} shows that SR is essentially flat for
$\alpha\!\in\![0.0,0.6]$ (within 0.6 points of the default 45.7\%)
and collapses to 39.3\% at $\alpha\!=\!1.0$, the same drop seen in
the \emph{text only} ablation---layout signal is doing the heavy
lifting.  Figure~\ref{fig:sens-tau} shows the $\beta$ sweep behaving
as a precision/recall trade-off: $\bar{\mathcal{N}}$ slides from
5.6 down to 4.7 as $\beta\!\to\!0$ while SR moves only between
45.3\% and 45.7\%, so $\beta$ is a deployment knob rather than a
critical hyperparameter.  Figure~\ref{fig:sens-backbone} swaps the
LLM backbone: Claude-4 gives 54.7\% SR and $\bar{\mathcal{N}}=5.1$ versus 45.7\% SR and $\bar{\mathcal{N}}=5.4$ on
GPT-4.1, a $+$9-point SR shift. The gains come from the retrieval
design and ride on top of whatever absolute performance the
backbone supplies.


\section{Conclusion}



We presented \SkillMigrator{}, a web agent that keeps the
standard \emph{text snapshot + tool calling} interface used by
recent skill-induction systems~\citep{wang2025inducing,yu2026polyskill,zheng2025skillweaver}
but indexes induced skills as \emph{transferable interaction
patterns (\TIP{})}---pairing each skill with a structural sketch of the
snapshot at induction time, and retrieving globally by layout
similarity before grounding to live element references.
On Mind2Web and WebArena, this reduces the average
LLM-action count on successful trajectories by 8--10\%
over the state-of-the-art methods, at matched task success rate.

\FloatBarrier

\bibliographystyle{unsrtnat}
\bibliography{neurips_refs}

\appendix

\section*{Appendix}
\section{Detailed Problem Formulation}
\label{app:formulation}

This appendix expands the description of the web-agent
setting that we summarised in \S\ref{sec:background}.

\paragraph{Web agent environment.}
We model web navigation as a partially observable sequential
decision process, following the standard view of interactive
agents in uncertain environments~\citep{kaelbling1998planning}.
A task is specified by a natural-language instruction $q$.  Since
a real web application may contain hidden browser state,
server-side state, and asynchronous page updates, the agent only
observes an agent-visible representation of the environment.
We use an observation-level abstraction and denote the
observation space by $\mathcal{S}$.  At time step $t$, the agent
receives an observation $o_t \in \mathcal{S}$, which may include
the rendered viewport, the DOM, the accessibility tree, and other
browser metadata.  This representation is consistent with prior
web-agent benchmarks, where agents interact with web pages
through visual and structural
observations~\citep{shi2017world,deng2023mind2web,zhou2024webarena,koh2024visualwebarena}.

Given the current observation, the agent selects an executable
action using an LLM $\pi_\theta$ via $a_t \sim
\pi_\theta(\cdot\mid q, o_{0:t}, a_{0:t-1})$, where $o_{0:t}$ and
$a_{0:t-1}$ are observation and action histories, respectively.
All actions are chosen from the valid action space $\mathcal{A}$
defined in Table~\ref{tab:action-space}, such as
\texttt{click(}$e$\texttt{)} and
\texttt{fill(}$e$,\,\textit{text}\texttt{)}.  After execution,
the web environment returns a new observation,
$o_{t+1} = \mathcal{T}(o_t, a_t)$, where $\mathcal{T}: \mathcal{S}
\times \mathcal{A} \to \mathcal{S}$ is the transition function
induced by the browser and the web application.  For simplicity,
we write $\mathcal{T}$ as deterministic, although real web
transitions may be stochastic due to network latency, dynamic
content, or nondeterministic server responses.

\paragraph{Subtasks and the LLM-action count.}
Given an instruction $q$, the planner decomposes it into
subtasks $\mathbf{s}(q)=\{s_1,\ldots,s_{T_q}\}$.  Let
$\tilde{o}_i$ denote the observation after finishing subtask
$s_i$, so $\tilde{o}_{i-1}$ is the observation before executing
$s_i$.  For each pair $(s_i,\tilde{o}_{i-1})$, the agent
retrieves a relevant skill from $\mathcal{K}$ when available.
If the subtask is fully covered by a retrieved skill, then
$\mathcal{N}(s_i,\tilde{o}_{i-1}\mid\mathcal{K},\pi_\theta)=0$;
otherwise the agent falls back to $\pi_\theta$ and generates a
trajectory $\tau = (o^{(0)},a_1,o^{(1)},\ldots,a_n,o^{(n)})$,
where $o^{(0)}=\tilde{o}_{i-1}$, $o^{(j)} =
\mathcal{T}(o^{(j-1)},a_j)$, and
$\mathcal{N}(s_i,\tilde{o}_{i-1}\mid\mathcal{K},\pi_\theta)=n$.
Equation~\eqref{eq:problem} of the main paper directly minimises the expectation
of $\sum_i \mathcal{N}(s_i,\tilde{o}_{i-1}\mid\mathcal{K},\pi_\theta)$
plus a library-cost regulariser $\lambda\,C(\mathcal{K})$.

\section{Experimental Setup Details}
\label{app:implementation}

This appendix expands the condensed Setup of
\S\ref{sec:exp:setup} with full benchmark protocol, baseline
regimes, and reproducibility details.

\paragraph{Benchmarks.}
\emph{Mind2Web}~\citep{deng2023mind2web} spans 137 websites across
31 domains and is partitioned into three generalisation splits:
cross-task (same website, new task), cross-website (same domain,
new website), and cross-domain (new domain).  Task success is
adjudicated by the previous GPT-4.1 judge protocol
\citep{yu2026polyskill, xue2025illusion}, which has $\sim$85\%
agreement with human judgments on this benchmark.
\emph{WebArena}~\citep{zhou2024webarena} provides 812 executable
tasks across shopping, shopping-admin, reddit, gitlab, map, and
multi-site columns, scored with the official programmatic
validators.  We follow the same accessibility-snapshot and
tool-calling protocol used by PolySkill, which keeps the action
schema fixed across compared systems and isolates the
contribution of skill retrieval and grounding.



\paragraph{Implementation.}
Our system is implemented on top of the Playwright MCP framework \footnote{\url{https://github.com/microsoft/playwright-mcp}} and AgentScope~\citep{agentscope_v1} for unified
agent orchestration and reasoning. In addition,
we use \texttt{sentence-transformers/all-MiniLM-L6-v2}
\citep{reimers2019sentence} as the frozen sentence encoder
$\mathbf{e}(\cdot)$, with no fine-tuning on \SkillMigrator{}
data.  All embedding similarities reported in Eq.~(3) are cosine
similarities of L2-normalised vectors.  Tree-edit distance for
the layout signal is computed by APTED~\citep{pawlik2016tree} on
the cleaned accessibility-tree skeletons.


\paragraph{Skill induction parity.}
Skills are induced from the same training trajectories used by
ASI~\citep{wang2025inducing} and
PolySkill~\citep{yu2026polyskill}, with their respective
induction pipelines, so that the only variable across systems is
the retrieval and grounding stack of \SkillMigrator{}.  All
reported numbers are means over three random seeds.

\section{Related Work}
\subsection{LLM-based Web Agents and Evaluation Environments}
Web agents aim to translate natural-language instructions into executable browser actions over interactive webpages. This problem has been studied through increasingly realistic benchmarks and environments, including WebShop~\cite{yao2022webshop}, Mind2Web~\cite{deng2023mind2web}, WebArena~\cite{zhou2024webarena}, and Branch-and-Browse~\cite{he2025branch}. In these settings, agents typically operate in a step-wise decision loop, observing the current webpage state and predicting one low-level action at a time, often following reasoning-and-acting paradigms such as ReAct~\cite{yao2023react}. This design provides broad flexibility, but it also makes deployment costly on long-horizon tasks, since each primitive action may require a separate policy-facing LLM call. Subsequent work further expanded the evaluation landscape to multimodal and realistic settings, such as WebVoyager~\cite{he2024webvoyager}, VisualWebArena~\cite{koh2024visualwebarena}, WorkArena~\cite{drouin2024workarena}, Odysseys~\cite{jang2026odysseys}, and MolmoWeb~\cite{gupta2026molmoweb}. Recent analysis has also highlighted that benchmark improvements do not always imply robust general web competence, motivating more careful measurement of efficiency and generalization in addition to task success~\cite{xue2025illusion}.

\subsection{Reusable Web Skills}
To reduce repeated low-level reasoning, a growing line of work studies reusable web skills, which compress frequently occurring primitive action sequences into higher-level abstractions. One line of work stores reusable procedural knowledge in textual form. Agent Workflow Memory (AWM)~\cite{wang2024agent} induces workflow memories from successful episodes and reuses them to guide future problem solving, improving long-horizon behavior through experience reuse. Another line of work emphasizes executable skill representations. ASI~\cite{wang2025inducing} induces programmatic skills from successful trajectories and exposes them directly as callable actions. SkillWeaver~\cite{zheng2025skillweaver} studies self-improvement through exploration, discovery, and iterative honing of reusable web skills. WALT~\cite{prabhu2025walt} further argues that agents should exploit higher-level website tools rather than rely entirely on brittle primitive UI interactions. PolySkill~\cite{yu2026polyskill} improves transfer by introducing polymorphic abstractions across websites, while more recent work explores richer skill formulations, such as combining executability with interpretability or improving repairability and verification~\cite{wang2026webxskill,lu2026contractskill}. Across these methods, the common goal is to shorten interaction horizons, reduce policy LLM calls, and improve robustness by reusing validated behavioral structure rather than regenerating every action from scratch.

Our work is most closely related to recent programmatic-skill web agents~\cite{wang2025inducing,zheng2025skillweaver,prabhu2025walt,yu2026polyskill}, but differs in what it treats as the main bottleneck. Rather than focusing primarily on how to induce or refine the internal implementation of a skill, we focus on how to retrieve and ground previously induced skills under broader transfer conditions. In particular, we target reuse beyond both the same website and the same domain. To do so, we index each skill not only by text but also by a structural sketch of the webpage snapshot at induction time, and retrieve from a global library using layout-aware matching before grounding abstract constraints to live element references. In this sense, our method complements prior work on workflow memory, executable skill induction, and polymorphic abstraction by emphasizing transferable interaction patterns as the retrieval key for cross-website and cross-domain reuse.

\section{Limitations}
\label{app:limitations}
Layout-conditioned retrieval assumes
that pages sharing accessibility-tree structure also share
interaction semantics. This holds for some
workflows we evaluate but weakens on visually isomorphic pages
whose control flow diverges, where lightweight multimodal
cues~\citep{he2024webvoyager,koh2024visualwebarena} would be a
natural complement to our text-only setting.
In addition, the operation templates are mined from observed action shapes, so modalities outside this distribution (e.g.\ drag-and-drop, modal dialogs) fall back to react mode until the inventory is extended. \SkillMigrator{} reduces the LLM-call cost of web automation, which can
improve efficiency and latency for repeated tasks. At the same time,
lower-cost automation may increase the scale of automated web actions.
Deployments should respect host-site usage policies, rate limits,
authentication boundaries, and user consent requirements, and should not
use skill reuse to bypass access controls or generate abusive traffic.

\end{document}